\documentclass{article}
\usepackage{graphicx, lineno, amsmath}
\usepackage[a4paper, margin=3cm]{geometry}
\usepackage{newtxtext,newtxmath}
\usepackage{setspace}

\usepackage{natbib}

\usepackage{siunitx}
\usepackage{booktabs}
\usepackage{adjustbox}
\usepackage[table]{xcolor}  %
\usepackage{subcaption}
\usepackage{hyperref}
\usepackage{xurl}
\usepackage{authblk}
\usepackage{multirow}
\usepackage{threeparttable}
\usepackage{array}
\usepackage{colortbl}
\usepackage{bbm}
\usepackage{svg}
\usepackage{makecell}
\usepackage{threeparttable}

\usepackage{verbatim}

\definecolor{lightgreen}{HTML}{B2DBA9}
\definecolor{lightorange}{HTML}{FDC872}
\definecolor{lightred}{HTML}{F0A59E}

\makeatletter
\renewcommand\maketitle{
  \begin{flushleft}
    {\LARGE\bfseries \@title \par}
    \vskip 1em
    {\large \@author \par}
  \end{flushleft}
}
\makeatother

\newcommand\DATASETNAME{SAGE}

\newcommand{\pms}[1]{{\,\scriptsize±\,#1}}

\title{SAGE: A sampling-aware global evaluation benchmark for species distribution modeling}

\author[1,a,*]{Emilia Arens}
\author[2,a]{Nina van Tiel}
\author[2,a]{Robin Zbinden}
\author[1]{Damien Robert}
\author[1]{Lukas Drees}
\author[2]{Chiara Vanalli}
\author[2]{Benjamin Kellenberger}
\author[3]{Niklaus E. Zimmermann}
\author[4,5]{Lo\"{\i}c Pellissier}
\author[2]{Devis Tuia}
\author[1]{Jan Dirk Wegner}

\affil[1]{EcoVision Lab, Department of Mathematical Modeling and Machine Learning, University of Zurich, Zurich, Switzerland}
\affil[2]{ECEO, École Polytechnique Fédérale de Lausanne, Sion, Switzerland}
\affil[3]{Swiss Federal Institute for Forest, Snow and Landscape Research, WSL Birmensdorf, Switzerland}
\affil[4]{Ecosystems and Landscape Evolution, Institute of Terrestrial Ecosystems, Department of Environmental Systems Science, ETH Zürich, Zürich, Switzerland}
\affil[5]{Land Change Science Research Unit, Swiss Federal Institute for Forest, Snow and Landscape Research, WSL Birmensdorf, Switzerland\vspace{0.5em}}
\affil[a]{Equal contribution}
\affil[*]{Corresponding author: emilia.arens@uzh.ch}

\date{} %

\begin{document}

\maketitle

\begin{abstract}
Knowing where species occur is fundamental for biodiversity research and conservation planning. Species distribution models (SDMs) are a key tool in this effort, linking observed species occurrences to environmental conditions to estimate their spatial distribution. However, the accuracy of the resulting maps varies with the underlying data and models, making it essential to know for which species they can be trusted. Recent deep-learning-based SDMs (``DeepSDMs'') can now jointly model thousands of species at a global scale, drawing on hundreds of millions of community-science occurrence records. At this scale, averaging performance across species can hide substantial species-level variability, particularly for rare species, which are often of greatest conservation concern. The underlying records, moreover, are strongly biased, geographically and taxonomically, making occurrence counts misleading. Accounting for these factors is therefore essential for a more reliable and informative evaluation of multi-species SDMs, but has so far not been attempted in a standardized way at large scales. Here, we introduce a Sampling-Aware Global Evaluation (\DATASETNAME) benchmark, combining occurrence records from the Global Biodiversity Information Facility (GBIF) for training with vegetation plots from sPlotOpen for presence-absence evaluation across \num{5771} plant species. We propose a sampling-aware evaluation framework built on two complementary species-level properties, sampling effort and relative prevalence, which describe how densely a species' range is sampled and how frequently the species itself is recorded. Grouping species by these properties reveals systematic performance differences across species and models that aggregate metrics cannot show. Evaluating both single-species SDMs and multi-species DeepSDMs, we find that Random Forests and DeepSDMs achieve the best overall performance, but neither approach dominates: DeepSDMs outperform single-species SDMs for infrequently recorded species while offering no consistent advantage for well-sampled ones. Crucially, this advantage emerges only when established bias-correction practices from the SDM literature, such as spatial thinning and reweighting, are carried over to the deep-learning setting. We demonstrate that \DATASETNAME{} helps identify the species and data conditions for which a given modeling approach is beneficial, thereby supporting the development of more transparent and ecologically credible species distribution models. Data, code, models and data visualizations are available at \url{https://earens.github.io/sage/}.

\end{abstract}

\vspace{0.5em}


\section{Introduction} 

Accurate mapping of species distributions is central to biodiversity research and conservation planning. Such maps support a wide range of conservation applications: prioritizing areas for protection~\citep{guisan2013predicting}, anticipating the impacts of climate change on biodiversity~\citep{franklin2023species,lawlor2024mechanisms}, managing invasive species~\citep{mainali2015projecting} and the spread of pests and pathogens~\citep{vaclavik2009invasive}, and planning infrastructure projects that minimize impacts on biodiversity~\mbox{\citep{baker2021species}}. However, species distribution maps are often imperfect estimates whose reliability depends on the models and data used to produce them, and their errors can propagate into the decisions they inform. Understanding when these maps can be trusted and how the models behind them behave is therefore essential.

\paragraph{Mapping species distributions.}
Species distribution maps can be generated using different methods. They range from expert-drawn maps to data-driven approaches that construct geographic polygons~\citep{natural2001iucn} or climate envelopes around known occurrence records~\citep{booth2014bioclim}. More flexible statistical approaches fit functions that link observed species occurrences to environmental conditions and project suitable habitat across space and time~\citep{guisan2005predicting, elith2009species}. This latter family, known as species distribution models (SDMs), has become the main methodological backbone for mapping species and is the focus of our work.

\paragraph{From single- to multi-species models.}
Traditionally, a separate SDM is fitted independently for each species~\citep{van2024regional} using established methods such as generalized linear models (GLMs), MaxEnt~\citep{phillips2006maximum}, or machine learning approaches such as Random Forests~\citep{breiman2001random, valavi2022predictive}. Species, however, do not occur in isolation. Joint species distribution models (JSDMs) account for this by modeling species together from community survey data, using patterns of co-occurrence to detect patterns indicative of species associations or biotic interactions~\citep{pollock2014understanding, ovaskainen2016uncovering}. Recent variants improve their scalability~\citep{tikhonov2020joint, pichler2021new} or extend them to occurrence records~\citep{molgora2022taxonomic}, but they have so far remained limited to at most a few hundred species. The increasing amount of available biodiversity data, together with the interest in scaling beyond a few hundred species, has accelerated the adoption of deep learning in ecology~\citep{pollock2025harnessing}, including multi-species deep learning SDMs (DeepSDMs) that jointly model many species through a shared representation~\citep{chen2017deep, Botella2018, deneu2021convolutional, zbinden2024selection, abdelwahed2026ciso}. DeepSDMs can scale to thousands of taxa and integrate heterogeneous predictors such as remote-sensing imagery, climatic time series, and textual species descriptions~\citep{teng2023satbird, picek2024geoplant,gillespie2024deep, dollinger2024sat, hamilton2024combining,zbinden2026miam}.

\paragraph{Data for species distribution modeling.}
All of these models are built on species occurrence data. The most reliable source is presence-absence (PA) data from systematic surveys, in which exhaustive species lists at each site record both detections and non-detections of species~\citep{liu2011measuring}. Such surveys are costly and labor-intensive and remain sparse at the global scale. Large-scale modeling therefore draws predominantly on presence-only (PO) data: occurrence records collected predominantly by volunteers through community-science platforms, which now provide hundreds of millions of observations using mobile devices and image-recognition tools~\citep{dickinson2010citizen,sullivan2014ebird,kosmala2016assessing}.

\begin{figure}
    \centering
    \begin{subfigure}[b]{\textwidth}
        \centering
        \includegraphics[width=\textwidth]{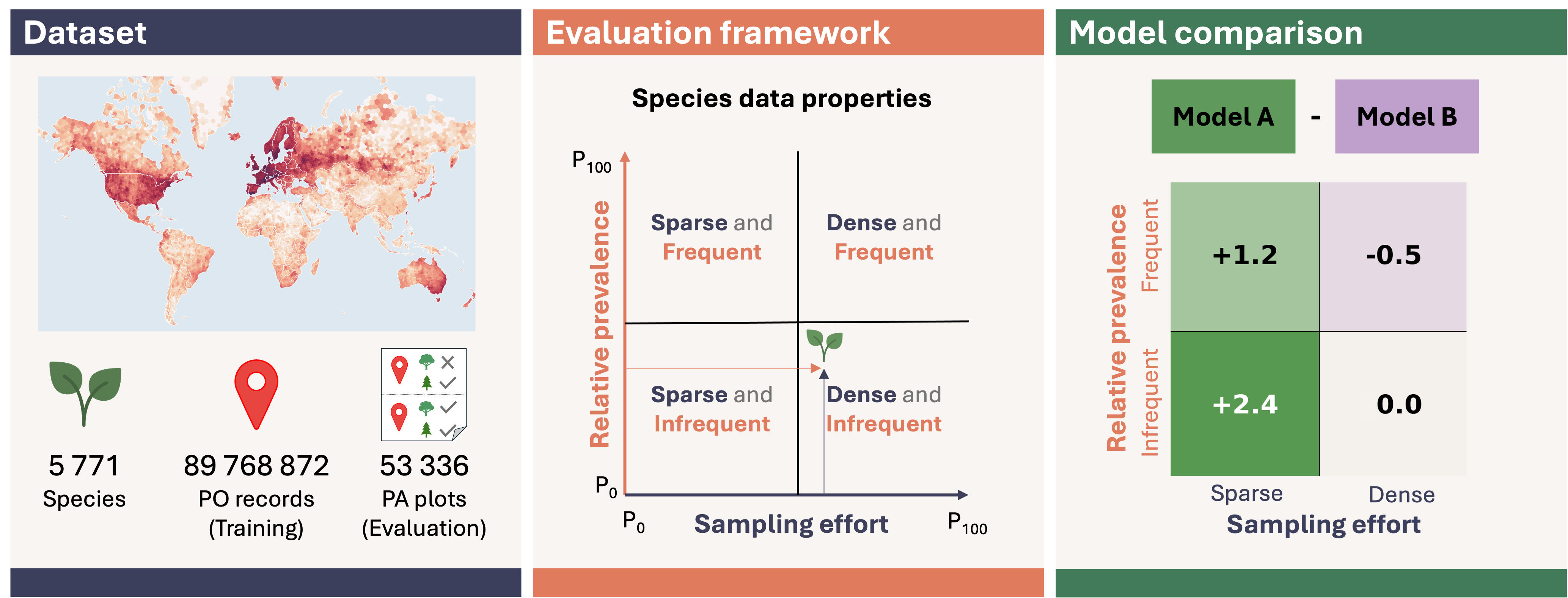}
    \end{subfigure}
    \caption{\textbf{Overview of \DATASETNAME{}.} \textbf{Dataset:} \num{89768872} presence-only (PO) occurrences from GBIF paired with \num{53336} expert-curated presence-absence (PA) vegetation plots from sPlotOpen \citep{sabatini2021splotopen}, covering \num{5771} non-anonymized plant species. \textbf{Evaluation framework:} each species is characterized by two properties, sampling effort and relative prevalence, defining a two-dimensional space partitioned into sparse versus dense sampling effort and infrequently versus frequently recorded species. \textbf{Model comparison:} models are compared across this space, with performance differences reported per quadrant rather than as a single average.}
    \label{fig:overview}
\end{figure}

\paragraph{Biases in community-science data.}
This abundance, however, comes at a price: community-science occurrence records exhibit multiple forms of bias \citep{bird2014statistical, callaghan2021three, johnston2023outstanding}. Spatial coverage is highly uneven, with greater sampling intensity in high-income regions and less than $7\%$ of the Earth’s $5$~km grid cells containing any occurrence record \citep{hughes2021sampling}. At finer scales, observations are concentrated in accessible or densely populated areas, leading to strong sampling gradients, for example, with increasing sampling density as distance to roads decreases \citep{kadmon2004effect, zizka2021sampbias, hughes2021sampling, geurts2023turning}. Taxonomic coverage is similarly skewed: some species are inherently harder to detect or identify, and observers tend to preferentially record conspicuous or charismatic taxa, leaving others underrepresented \citep{troudet2017taxonomic, callaghan2021three, geurts2023turning}. These biases propagate through the modeling pipeline and are often most pronounced for the underrepresented species, many of which are of greatest conservation concern. %

\paragraph{Evaluating multi-species models.}
Because conservation decisions depend on accurate predictions, knowing which model can accurately model the distribution of which species is essential. However, the shift toward large-scale, multi-species modeling introduces two challenges for evaluation. First, performance is typically computed for each species and then aggregated by averaging across species. While convenient, such summaries hide substantial variability and conceal which species each model performs best or worst on. At the same time, inspecting performance at the individual level becomes infeasible when the number of modeled species grows into the thousands. Second, a model's measured performance is entangled with the data it was trained on: its score reflects the conditions under which a species was sampled as much as the model's skill. Without accounting for these conditions, apparent differences between models may reflect the data rather than the models themselves, leaving the picture of model quality incomplete and potentially misleading.

\paragraph{Towards a sampling-aware evaluation.} Addressing these challenges requires looking beyond average performance to ask which species a model actually serves: does a model that scores well on average also map data-deficient species well or only common, well-recorded ones? Answering this question calls for an evaluation framework with two components: (i) a large-scale dataset that pairs presence-only training data with presence-absence evaluation; (ii) an evaluation protocol that goes beyond aggregate metrics by relating performance to the data conditions under which each species is observed and reporting it in a structured, disaggregated way.

\definecolor{cellgreen}{HTML}{A8D5A2}
\definecolor{cellorange}{HTML}{F5C87A}
\definecolor{cellred}{HTML}{E8A09A}

\newcommand{\cbar}[2]{\raisebox{-5pt}{\parbox[b]{1.9cm}{\rule{0pt}{1.2em}\centering #1\par\nointerlineskip\vspace{2pt}\textcolor{#2}{\rule{1.9cm}{4pt}}}}}
\newcommand{\cellg}[1]{\cbar{#1}{cellgreen}}
\newcommand{\cello}[1]{\cbar{#1}{cellorange}}
\newcommand{\cellr}[1]{\cbar{#1}{cellred}}

\begin{table}
    \caption{\textbf{Comparison of existing SDM benchmark datasets.} Color indicators reflect the degree to which each criterion supports transparent, species-resolved evaluation: \textcolor{cellgreen}{\rule{1em}{3pt}}~fully, \textcolor{cellorange}{\rule{1em}{3pt}}~partially, \textcolor{cellred}{\rule{1em}{3pt}}~not fulfilled. \textsuperscript{\dag}Training data only; evaluation covers up to \num{2418} species.}
    \centering
    \renewcommand{\arraystretch}{1.3}
    \setlength{\arrayrulewidth}{0.5pt}
    \begin{tabular}{b{3cm} cccc}
        \toprule
        \textbf{Benchmark}
        & \textbf{Geographic Scope}
        & \textbf{\# Species}
        & \makecell{\textbf{Species}\\\textbf{Identity}}
        & \textbf{Evaluation Data} \\
        \midrule

        \cite{elith2020presence}
        & \cello{Regional}
        & \cello{\num{223}}
        & \cellr{No}
        & \cellg{Plot data} \\
        \hline

        \cite{cole2023spatial}
        & \cellg{Global}
        & \cellg{\num{47375}\textsuperscript{\dag}}
        & \cellg{Yes}
        & \cellr{Expert ranges} \\
        \hline

        SatBird \citep{teng2023satbird}
        & \cello{Regional}
        & \cello{\num{670}}
        & \cellg{Yes}
        & \cello{Community science} \\
        \hline

        GeoPlant \citep{picek2024geoplant}
        & \cello{Europe}
        & \cellg{\num{5016}}
        & \cellr{No}
        & \cellg{Plot data} \\

        \midrule[0.12em]

        \DATASETNAME\ (ours)
        & \cellg{Global}
        & \cellg{\num{5771}}
        & \cellg{Yes}
        & \cellg{Plot data} \\

        \bottomrule[0.12em]
    \end{tabular}
    \label{tab:benchmark}
\end{table}

\paragraph{Existing benchmarks.}
These needs are not new: the presence-only-training, presence-absence-evaluation paradigm was established by early benchmarking efforts, most influentially the NCEAS working group and the public benchmark of \cite{elith2020presence}. This benchmark, however, is based on relatively small, regional datasets and anonymized species identities, making it too limited to evaluate the large-scale, multi-species models now common \citep{zbinden2024selection}, and preventing performance from being related to specific species and linked to broader species knowledge beyond the data itself. 
More recent initiatives expand coverage, but each falls short on at least one aspect (Table~\ref{tab:benchmark}):~\cite{cole2023spatial} includes more than \num{47000} species worldwide, but evaluates only on a subset of them and uses coarse expert range maps rather than observation record data. SatBird~\citep{teng2023satbird} provides eBird\footnote{\url{https://ebird.org}} checklists for training and evaluation, but remains geographically and taxonomically narrow. GeoPlant~\citep{picek2024geoplant} offers European plot-based evaluation, but anonymizes species. No existing benchmark simultaneously provides global coverage, a large number of non-anonymized species, and reliable plot-based presence-absence evaluation. Additionally, no benchmark explicitly accounts for species-level differences in data quality or groups species by key properties, such as sampling effort or prevalence, when evaluating model performance.

\paragraph{The \DATASETNAME{} benchmark.}
We address these limitations with the Sampling-Aware Global Evaluation benchmark (\DATASETNAME{}, Figure~\ref{fig:overview}), a dataset and evaluation framework that extends the presence-only-training, presence-absence-evaluation paradigm of~\cite{elith2020presence} to large-scale, multi-species modeling. \textit{First}, we release a curated global plant dataset that pairs \num{89768872} citizen-science occurrences from the Global Biodiversity Information Facility (GBIF) for training~\citep{gbif2026download} with \num{53336} expert-curated sPlotOpen vegetation plots for presence-absence evaluation~\citep{sabatini2021splotopen}, covering \num{5771} non-anonymized species. Each sample is paired with \num{52} tabular environmental predictors and each species with its POWO native range \citep{powo2025}; the training data is additionally provided in aggregated and subsampled versions that apply standard data-thinning strategies to mitigate sampling bias and reduce computational cost~\citep{boria2014spatial, sillero2021want}. 
\textit{Second}, we characterize each species by two properties, sampling effort and relative prevalence, that serve as proxies for how thoroughly its range is sampled and how often it is recorded where sampling occurs. By grouping species according to these properties and analyzing how model performance varies across them, we provide a more transparent and ecologically meaningful assessment of where models perform well and where they fail. \textit{Third}, we evaluate both single-species SDMs and multi-species DeepSDMs on \DATASETNAME. We find that no single approach dominates: Random Forests and DeepSDMs perform comparably overall, yet their strengths diverge across species. DeepSDMs are most beneficial for infrequently recorded species, but this advantage emerges only when data preparation and the training objective are tailored to the multi-species setting, drawing on established bias-correction practices such as spatial thinning and reweighting.

Together, these contributions support the development of more transparent, reliable, and ecologically credible multi-species distribution models, with the aim of improving biodiversity monitoring.

\section{Materials and Methods}
\subsection{Datasets}
\label{sec:datasets}
\begin{figure}
    \centering
    \begin{subfigure}[b]{0.99\textwidth}
        \centering
        \includegraphics[width=\textwidth]{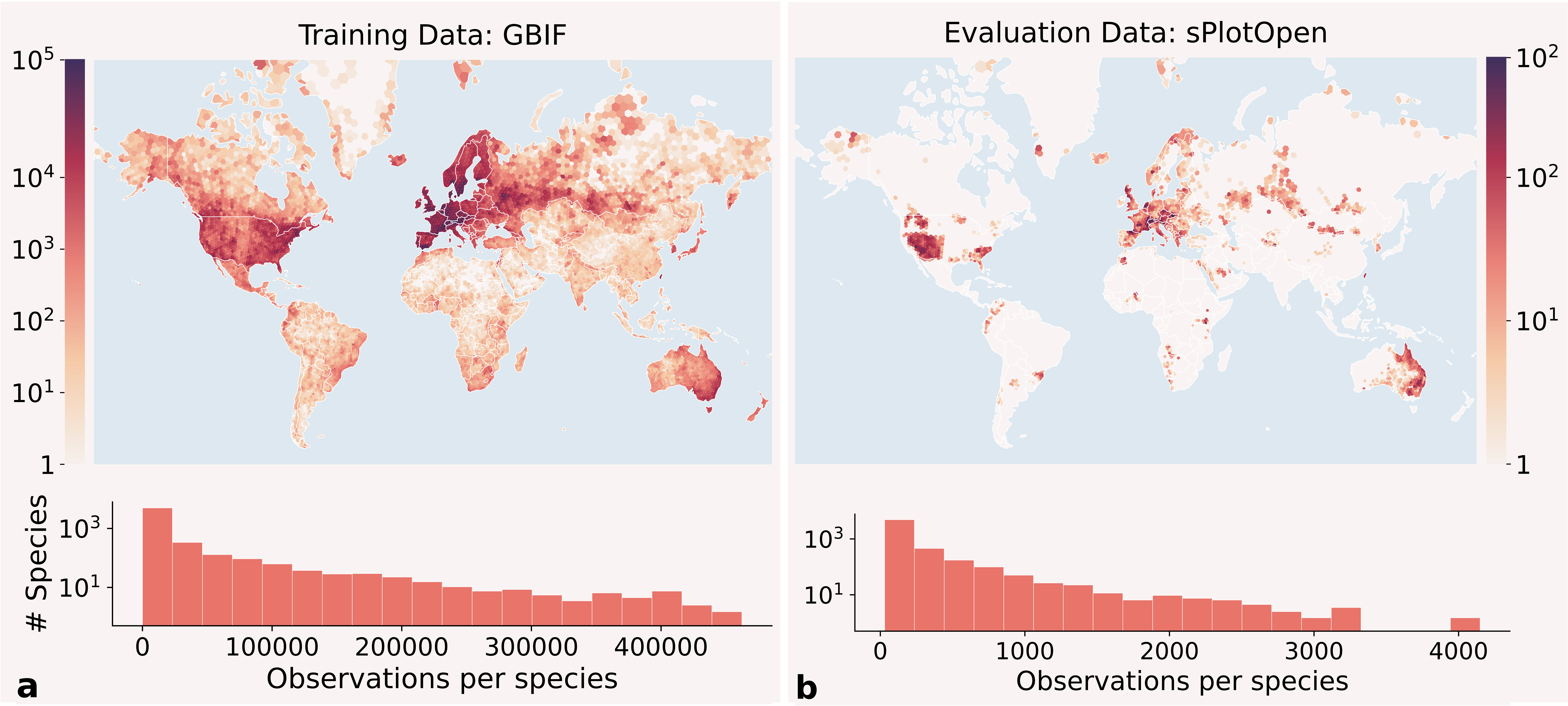}
    \end{subfigure}
    \caption{\textbf{Dataset statistics.} \textbf{(a)} Presence-only training data from GBIF (full variant: \num{89768872} occurrence records for \num{5771} species) and \textbf{(b)} presence-absence evaluation data from sPlotOpen vegetation plots (\num{53336} plots for \num{5771} species). In each panel, the map shows the geographic distribution of records, colored by occurrence density (log scale), and the histogram below shows the distribution of per-species occurrence counts. Color scales and histogram axes differ between panels, reflecting the much larger volume of GBIF data.}
    \label{fig:datasets}   
\end{figure}

\DATASETNAME{} is designed to meet two key objectives: leveraging the vast coverage of species observations offered by community-science data, and providing a reliable evaluation set with non-anonymized species information at the largest possible scale. This combination enables the systematic evaluation of SDMs, under realistic global-scale conditions (Table~\ref{tab:benchmark}). In line with the well-established benchmark of \cite{elith2006novel}, we provide presence-only data for model training and presence-absence data for model evaluation on the same \num{5771} plant species (Figure \ref{fig:datasets}). 
For each observation in our datasets, we provide a set of \num{52} environmental predictors commonly used for modeling purposes. The following subsections describe our datasets in more detail.

\subsubsection{Evaluation Data --- sPlotOpen Presence-Absence Records}
The sPlotOpen dataset \citep{sabatini2021splotopen} forms the foundation of our evaluation framework. 
To the best of our knowledge, it is the largest publicly available, globally distributed vegetation-plot dataset with non-anonymized species names.
The dataset provides systematically collected, expert-curated vegetation-plot observations with explicit species lists and detailed plot metadata.
sPlotOpen is an open-access subset of the larger sPlot dataset~\citep{bruelheide2019splot}, resampled to achieve a more environmentally balanced global distribution, resulting in 105 regional vegetation datasets and \num{95104} plots spanning \num{42677} vascular plant taxa, 114 countries, and all major terrestrial biomes.
Even after this environmental resampling, residual imbalance remains at the biome level, with temperate and mid-latitude ecosystems (e.g., in Europe, Australia and the USA) more densely sampled than tropical and subtropical biomes (Figure \ref{fig:datasets}b). Plot sizes vary widely, ranging between 0.03 and \num{40000} m$^2$, with a median of 100 m$^2$, and are often much smaller than the 1~km$^2$ resolution considered in our modeling pipeline, affecting the comparability of fine-grained plot data with coarser-scale model predictions. We quantify the effect of this mismatch on the evaluation in Appendix~\ref{sec:appendix_plot_area}, and find that it shifts absolute scores but leaves the model ranking unchanged.
It remains nonetheless the most readily usable global resource for presence-absence data, offering a uniquely broad taxonomic and geographic coverage for structured model evaluation.

To prepare our evaluation dataset, we exclude plots with reported coordinate uncertainty exceeding 1~km to match the spatial resolution of our predictors~\citep{moudry2024optimising}, and remove plots lacking complete species inventories to ensure reliable absence records. We merge surveys that occur within the same 1~km grid cell: we follow \cite{elith2006novel} and treat a species as present if it is reported in any contributing record, even if another survey in the same cell did not detect it. We note that we merge surveys based on geographic location rather than plot identifiers, thereby collapsing the temporal axis.
Taxonomic names are harmonized with the World Checklist of Vascular Plants (WCVP) \citep{govaerts2021world} using fuzzy matching and synonym resolution to guarantee consistent species identities across all data sources. To ensure robust species-level performance estimation, we retain only taxa with at least 30 occurrences in the cleaned evaluation set. The curated output results in presence-absence data for \num{53336} plots containing \num{922572} observation records for \num{5771} species (Figure \ref{fig:datasets}b).

The evaluation dataset is split into 20\% validation (\num{11068} plots) and 80\% test set (\num{42268} plots). To ensure reliable per-species evaluation, the split is computed using an iterative stratification procedure that processes species from rarest to most common and reserves enough plots for each species so that every species has at least 20 occurrences in the test set \citep{elith2020presence}.
We tune model hyperparameters on the validation set before evaluating on the test set, which is used solely to assess predictive performance. Maintaining separate validation and test sets is standard practice in machine learning to prevent information from the model-selection process from influencing the final performance estimate and thus provides a more reliable assessment of generalization \citep{james2013introduction}.

Additional details of the data curation workflow are provided in Appendix~\ref{sec:appendix_splotopen}, and more detailed dataset statistics are presented in Appendix~\ref{sec:appendix_data_stats}.

\subsubsection{Training Data --- GBIF Presence-Only Records}
\label{sec:PO_data}

To obtain a global presence-only training dataset, we extract all GBIF \citep{gbif2026download} records for the species recorded in the evaluation set. We query GBIF for georeferenced occurrences, download the complete record set, and harmonize nomenclature using the same WCVP resolution pipeline. Records flagged by the GBIF coordinate quality filters or by standard coordinate-cleaning heuristics (e.g., zero coordinates, country centroids, city proximity) are excluded. Occurrence records with identical coordinates are merged, creating co-occurrence records and removing duplicates. We do not, however, exclude occurrences that fall within evaluation cells, as the presence-only training and presence-absence evaluation data come from independent sources with markedly different distributions. This setting applies identically to all evaluated models. The detailed cleaning pipeline and parameters can be found in Appendix~\ref{sec:appendix_gbif}.

We exclude occurrences falling outside the corresponding species' range: these ranges are constructed from reported native country checklists from Plants of the World Online (POWO) \citep{powo2025} and country polygons from \citet{brummitt2001world}. This restricts the data to regions where species are reported as native, minimizing the effects of including habitat suitability in areas where the species is introduced or invasive \citep{van2024regional, enquist2026bien}. We note that for some larger countries, checklists and polygons were available at a sub-country regional level. Details on range construction can be found in Appendix~\ref{sec:appendix_range_maps}.

To evaluate the effect of different bias-correction methods, we construct three versions of the training dataset, reflecting common spatial and sampling standardization strategies \citep{boria2014spatial, sillero2021want}:
\begin{itemize}
    \item \textbf{Full.} The complete presence-only dataset produced by the above-described procedure contains \num{89768872} occurrence records for \num{5771} species across \num{30138089} unique locations. The data at each location is considered a sample, defined by unique coordinates, which may contain records for one or more species (Figure \ref{fig:datasets}a).
    \item \textbf{Aggregated.} In the spatially aggregated version, all presences within a 1~km cell are merged, reducing spatial autocorrelation introduced by clustered sampling at accessible locations \citep{boria2014spatial}. A species is considered present in a cell (i.e., a sample) if at least one record for that species falls within it.  Aggregation also facilitates multi-species learning by incorporating local co-occurrence structure while reducing storage and computational cost. Because most predictors used here have a 1~km resolution, their values are directly associated with the corresponding grid cells; for higher-resolution environmental predictors, we use the average value within each cell. This version contains \num{57084622} occurrence records, a \num{36.4}\% reduction compared to the full dataset, across \num{4100969} samples, a \num{86.4}\% reduction, while retaining all \num{5771} species.
    \item \textbf{Subsampled.} This version balances species counts by randomly sampling a fixed number of samples per species from the aggregated dataset (or retaining all samples for species with fewer samples than the threshold). Because each sample records all co-occurring species, a sample selected for one species also contributes observations for any other species present there. As a result, species may end up with more records than the predefined target. This version attenuates the long-tailed distribution by reducing the disparity in the number of occurrences across species, while further reducing storage and computational cost through a smaller number of samples. Different subsampling thresholds were considered, but our main subsampled dataset applies a threshold of \num{1000} samples per species, resulting in \num{43071245} occurrence records, a \num{52.0}\% reduction compared to the full dataset, across \num{1627928} samples, a \num{94.6}\% reduction, while retaining all \num{5771} species.
\end{itemize}

\subsubsection{Pseudo-absences}

Since our training data is presence-only, we sample pseudo-absences to contrast observed presences when training models \citep{pearce2006modelling, barbet2012selecting}. We consider two types of pseudo-absences commonly used in the literature: random background points and target-group background points. Random background points are generated using a Fibonacci lattice to distribute points regularly over the spherical Earth \citep{russwurm2024locationencoding}. We set the spacing between points to approximately 5~km, resulting in a total of \num{5865306} random background points after removing those that do not fall on land. In addition, we use target-group background points, i.e., treating the presence of other species as absence points. These are readily available because our dataset includes multiple species. Target-group background points are known to help mitigate geographic biases in presence-only data by providing the model with information about the overall sampling effort across space \citep{phillips2009sample, botella2020bias}. 

\subsubsection{Predictors}

For each sample in our training, validation, and test sets, we collect environmental predictors to characterize the abiotic conditions at the given location. We focus on predictors commonly used in SDMs that are linked to plant suitability \citep{mod2016we, fourcade2018paintings}. Specifically, we include the 19 bioclimatic predictors related to temperature and precipitation from CHELSA v2.1 at 1~km resolution \citep{brun2022chelsa}, 8 soil property variables from SoilGrids at 250~m resolution \citep{hengl2017soilgrids250m}, and 16 topographic variables (including elevation, slope, and aspect) obtained from \citet{amatulli2018suite} at 1~km resolution. Finally, we account for anthropogenic influence by incorporating 9 human footprint variables from \cite{venter2016global}, which describe human pressures such as the built environment and transportation infrastructure at 1~km resolution. In total, this results in \num{52} raster predictors. The complete list of predictors used is provided in Appendix \ref{sec:appendix_predictors}. For each sample, we extract a single scalar value for each predictor from the corresponding grid cell, resulting in a tabular data format.

\subsection{Evaluation Framework}
\label{sec:evaluation}

We evaluate models trained on presence-only data by assessing predictive performance on our presence-absence test dataset. Model performance is evaluated at the species level using the Area Under the Receiver Operating Characteristic curve (AUROC) and the Area Under the Precision-Recall-Gain curve (AUPRG; \citealp{flach2015precision}, Appendix \ref{sec:appendix_auprg}). For each species, we compute these metrics only within its native range: scoring over the entire globe would count vast regions where a species cannot occur as trivially correct absences and thereby inflate the threshold-independent metrics \citep{lobo2008auc, barve2011crucial}. This procedure also excludes locations that may be suitable but fall outside the native range, whether unoccupied due to dispersal limitation or historical factors, or occupied due to invasive/naturalized populations. Disentangling these dynamics is beyond our scope, so we restrict both training and evaluation data to each species' native range, excluding such records entirely.

As discussed in the introduction, evaluating thousands of species jointly makes it intractable to diagnose performance taxon by taxon, while averaging performance scores across all species may hide interesting patterns, such as the influence of data biases. 

\begin{figure}
    \centering
    \begin{subfigure}[b]{0.99\textwidth}
        \centering
        \includegraphics[width=\textwidth]{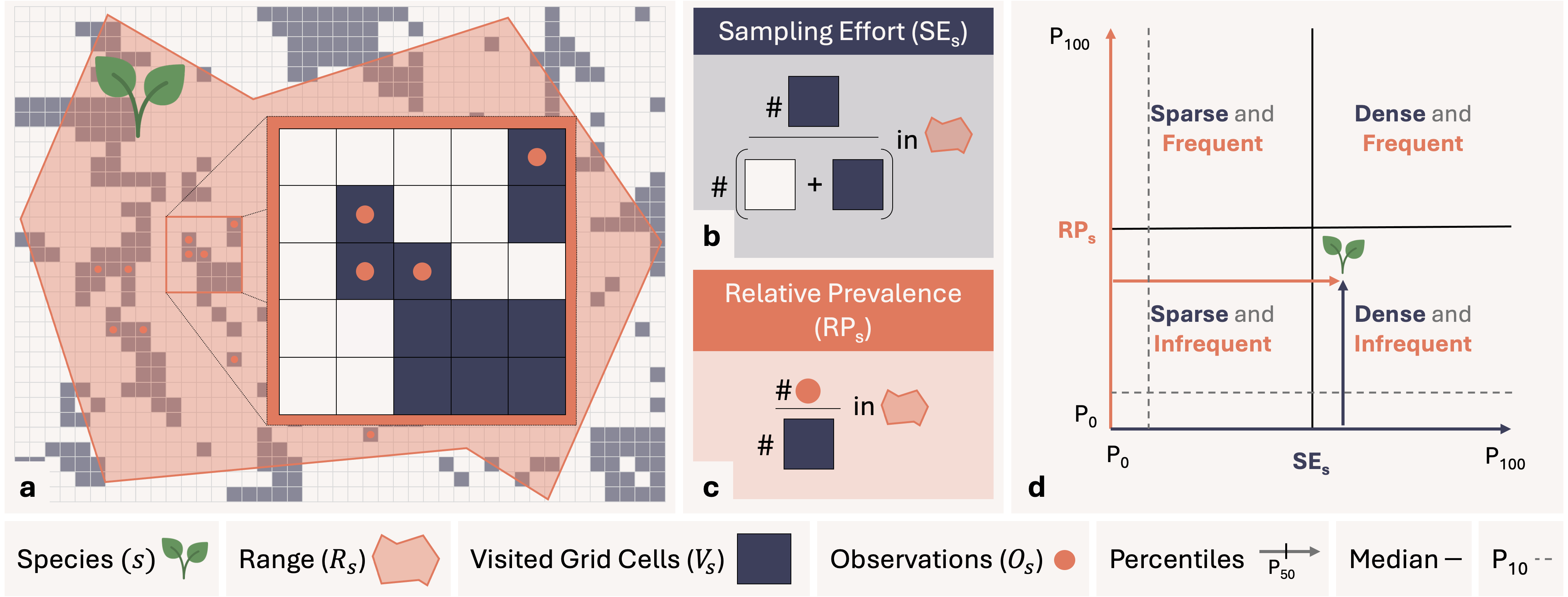}
    \end{subfigure}
    \caption{\textbf{Conceptual illustration of species data properties.} From the distribution of the observations in range \textbf{(a)}, we derive two complementary properties. \textbf{(b) Sampling Effort} ($\text{SE}_s$): the fraction of grid cells within a species' range that contain at least one occurrence record (from any species). \textbf{(c) Relative Prevalence} ($\text{RP}_s$): the fraction of visited cells within the species' range in which the species was recorded. \textbf{(d) Property Space}: the resulting two-dimensional space, where species are grouped into four quadrants defined by the median of each property. Species in the lowest decile of either property (i.e., the sparsest or most infrequent taxa; ``$\mathrm{P_{10}}$'') are also highlighted.}
    \label{fig:properties_conceptual}
\end{figure}

To obtain a structured and interpretable characterization of the variability of occurrence data across species and how that may influence model behavior, we consider two properties for each species: sampling effort and relative prevalence. These species-specific properties capture complementary aspects of data availability, reflecting both sampling coverage and species detectability within sampled locations. They are computed for each species using the presence-only training data after aggregation to the 1~km pixel grid, with the species' range serving as the reference domain. Figure \ref{fig:properties_conceptual} illustrates these properties.

\paragraph{Sampling effort.}
We define the sampling effort SE$_s$ for species $s$ as the proportion of grid cells within the range of $s$ that contain at least one occurrence record, regardless of whether species $s$ was observed:
\[
\text{SE}_s = \frac{|V_s|}{|R_s|},
\]
where $R_s$ denotes the set of grid cells in the expert range of species $s$, and $V_s \subseteq R_s$ refers to the subset of grid cells that contain at least one occurrence record (from any species) in the presence-only training data. We underline that we consider records independently of species identity, thereby quantifying the overall sampling effort within the species' range rather than the species-specific sampling effort. Consequently, species with identical ranges also have identical sampling effort values.
Intuitively, SE$_s$ measures how well the range of species $s$ has been sampled. It therefore enables comparisons of sampling intensity across species ranges and captures geographic variation in the coverage of community-science observations.

\paragraph{Relative prevalence.}
We define the relative prevalence RP$_s$ of species $s$ as the number of grid cells in which there is a record of $s$ divided by the total number of visited grid cells within the expert range of $s$:
\[
\text{RP}_s = \frac{|O_s|}{|V_s|},
\]
where $O_s \subseteq V_s$ denotes the subset of grid cells in which species $s$ was observed. Because these properties are computed after aggregating occurrence data to the grid-cell level, the number of grid cells in which $s$ was observed is equivalent to the number of occurrences of $s$ in the aggregated dataset.
In other words, RP$_s$ captures how frequently a species is observed within the sampled portion of its range. Low relative prevalence may reflect at least two distinct phenomena that cannot be disentangled from occurrence records alone: taxonomic sampling bias (e.g., observers preferentially record edelweiss rather than grass) or genuinely low species abundance (e.g., edelweiss are intrinsically rare).

\begin{figure}
    \centering
    \begin{subfigure}[b]{0.99\textwidth}
        \centering
        \includegraphics[width=\textwidth]{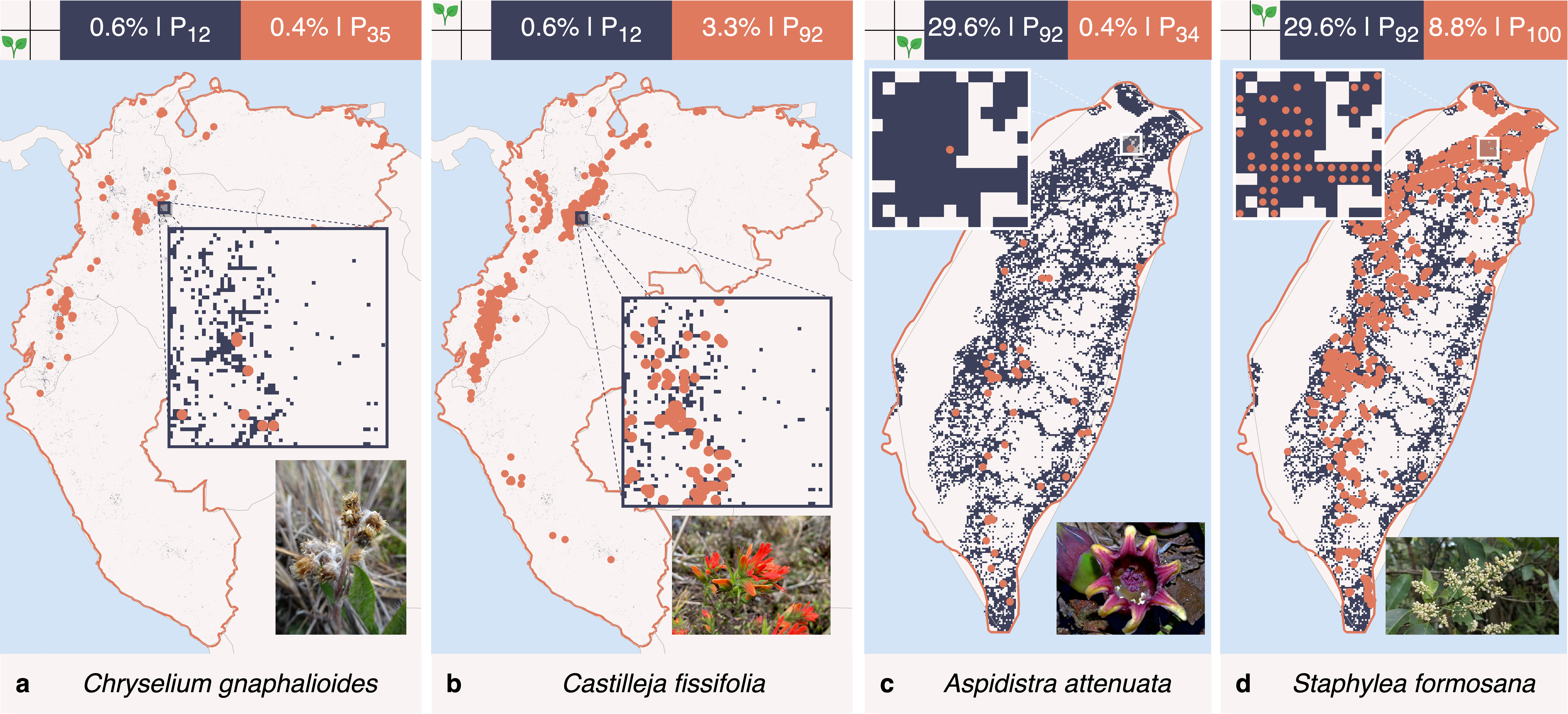}
    \end{subfigure}
    \caption{\textbf{Examples of species representing the four quadrants of the property space}, with their sampling effort (SE in blue) and relative prevalence (RP in orange), and their corresponding percentile ranking (P): \textbf{(a)} \textit{Chryselium gnaphalioides} sparse and infrequent; \textbf{(b)} \textit{Castilleja fissifolia} sparse and frequent; \textbf{(c)} \textit{Aspidistra attenuata} dense and infrequent; \textbf{(d)} \textit{Staphylea formosana} dense and frequent. Species photos: \textit{Chryselium gnaphalioides} \copyright~Fabien Anthelme, Pl@ntNet, CC BY-SA 2.0 (observed April 3, 2019), \textit{Castilleja fissifolia} \copyright~Fabien Anthelme, Pl@ntNet, CC BY-SA 2.0 (observed November 25, 2021), \textit{Aspidistra attenuata} \copyright~Jacy Chen, iNaturalist, CC BY 4.0 (observed December 9, 2021), \textit{Staphylea formosana} \copyright~Cheng-Te Hsu, iNaturalist, CC BY 4.0 (observed March 31, 2026).}
    \label{fig:properties_maps}
\end{figure}

\paragraph{Relationships between the properties and the number of occurrences.}

After aggregation to the grid-cell level, the number of occurrences of a species $s$ can be expressed as the product of sampling effort, relative prevalence, and the range size:
\[
|O_s| = \frac{|V_s|}{|R_s|} \cdot \frac{|O_s|}{|V_s|} \cdot |R_s| = \text{SE}_s \cdot \text{RP}_s \cdot |R_s|.
\]
This factorization highlights that a low number of occurrences may arise because any of these three factors is small, or because several of them are simultaneously small.

\paragraph{Species Groups in the Property Space.}
Rather than aggregating the predictive performance of all species together, we investigate how model performance varies according to the species data properties. Specifically, we consider a two-dimensional space defined by sampling effort and relative prevalence in which each species is positioned according to its values for the two properties (Figure \ref{fig:properties_conceptual}d). We divide species into groups at the median of each property, resulting in four groups corresponding to the four quadrants of the resulting two-dimensional property space. We refer to species with sampling effort lower than the median as having \textit{sparse} sampling, as opposed to \textit{dense} sampling for species with sampling effort above the median. Similarly, species with relative prevalence in the lower or upper half of values are referred to as having \textit{infrequent} or \textit{frequent} sampling, respectively. Figure \ref{fig:properties_maps} illustrates the observations and property values for one representative species from each quadrant. Species assigned to the \textit{sparse and infrequent} quadrant represent the most data-deficient cases, occurring in relatively poorly sampled regions and rarely recorded even where sampling exists. \textit{Sparse and frequent} species are frequently recorded where observers are active, but their ranges lack sampling coverage. Conversely, \textit{dense and infrequent} species inhabit well-sampled regions yet remain underrepresented in the observations. Finally, \textit{dense and frequent} species are the most data-rich, benefiting from both extensive sampling coverage and high recording rates. 
Beyond these four quadrants, we further distinguish species in the lowest decile of each property, split by the median of the other, yielding four additional groups that capture the species most affected by bias or rarity along each axis (Figure \ref{fig:properties_conceptual}d).

More information and visualizations of the proposed species data properties and the associated property space are provided in Appendix~\ref{sec:appendix_properties}.

\subsection{Models}
We compare the performance of a variety of models on \DATASETNAME. As our primary objective is to establish a large-scale benchmark for species distribution modeling, model selection is driven by the characteristics of the dataset rather than by a particular modeling paradigm: we restrict our study to approaches that can be trained on presence-only data and scale to \num{5771} species and millions of occurrences. For single-species SDMs, we evaluate well-established approaches such as MaxEnt~\citep{phillips2006maximum} and Random Forest~\citep{breiman2001random}. For multi-species modeling, the conventional JSDMs framework does not meet these requirements. JSDMs are typically designed for presence-absence data and, even in scalable or presence-only formulations, have been demonstrated only at much smaller scales~\citep{tikhonov2020joint, pichler2021new, molgora2022taxonomic}. We therefore focus on multi-species deep learning models (DeepSDMs). 

\subsubsection{Single-Species Models}

We benchmark established single-species SDM approaches, in which a separate model is fitted for each species, selecting methods that performed well on the \cite{elith2020presence} data \citep{valavi2022predictive}: a Generalized Linear Model (GLM), a Generalized Additive Model (GAM), MaxEnt, Boosted Regression Trees (BRT), and Random Forests (RF). Each model is trained on the aggregated dataset (Section~\ref{sec:datasets}), which is free of within-cell pseudo-replication. We cap presences at \num{10000} per species, which keeps training tractable across the \num{5771} individually fitted models and is well above the sample sizes on which these methods are typically developed and validated \citep{elith2020presence, wisz2008effects}. Presences are contrasted with up to \num{10000} target-group background points, which mitigate sampling bias by reflecting overall survey effort \citep{phillips2009sample, botella2020bias}, and the resulting class imbalance is handled by reweighting (GLM, GAM, BRT) or per-tree downsampling (RF) \citep{barbet2012selecting, valavi2022predictive}. Hyperparameters follow \citet{valavi2022predictive}, with each model's most important hyperparameter additionally tuned per species on the validation set, and the selected configuration retrained over five different random seeds. More details on the models and hyperparameters are provided in Appendix \ref{sec:appendix_single_species}.

\subsubsection{Multi-Species Deep Learning Models}
We consider DeepSDMs that jointly model all species through a shared representation. As a comparatively nascent approach in species distribution modeling, deep learning lacks the accumulated guidance available for established single-species methods and so requires more deliberate design choices. We therefore adapt the data-preparation and bias-correction practices established for the single-species models (aggregation, target-group background, and class rebalancing) to the joint setting. Starting from a baseline configuration, we introduce four design choices and isolate the contribution of each: two concern the training data (aggregation and subsampling) and two the model and its training (architecture and the loss). We describe each configuration below, highlighting the changes introduced by each design choice. Full details on the models and training procedures are provided in Appendix \ref{sec:appendix_multi_species}.

\begin{itemize}
    \item \textbf{Baseline.} Our baseline uses a simple four-layer multi-layer perceptron (MLP) with a hidden dimension of \num{512}. The model is trained on the full dataset using the AdamW optimizer \citep{loshchilov2017decoupled} and a binary cross-entropy loss combining random and target-group background points, following \cite{cole2023spatial} (Appendix~\ref{sec:appendix_loss}). 
    \item \textbf{Aggregation.} We replace the full dataset with its aggregated counterpart, the same data used for the single-species models, merging records within each 1~km cell. Beyond removing within-cell pseudo-replication, this resolves a conflict specific to joint training: without aggregation, a presence of one species and a target-group background point contributed by another can fall in the same cell, presenting the model with identical predictors but contradictory labels. 
    \item \textbf{Subsampling.} We then subsample the aggregated data, limiting the number of occurrences per species by using the procedure described in Section \ref{sec:PO_data}. Because multi-species models optimize all species jointly, this reduces the influence of data-rich taxa on the shared objective while reducing computational costs. We evaluate several thresholds and adopt \num{1000} occurrences per species for the main results and subsequent design choices.
    \item \textbf{Architecture.} We replace the baseline MLP with a residual network (ResNet) and an FT-Transformer, both adapted for tabular data \citep{gorishniy2021revisiting}, varying the hidden dimension and the number of residual or transformer blocks. Exploring this range ensures the deep models are evaluated at an appropriate capacity rather than constrained by an arbitrary default. For the main results, we use the ResNet with a hidden dimension size of \num{1024} and \num{8} blocks.
    \item \textbf{Loss.} Finally, we replace the baseline loss function with a weighted formulation \citep{zbinden2024selection}. It rebalances presences against pseudo-absences, as the single-species models do, and adds species-specific weights to counter imbalance across species. It also controls the relative contribution of random and target-group background points. For the main results, we use frequency-based weighting with target-group background only. Full definitions and the hyperparameters varied are provided in Appendix~\ref{sec:appendix_loss}.
\end{itemize}
We evaluate models with each of these design choices incrementally applied in the order in which they are listed here and refer to the model applying all design choices as the ``optimized deepSDM''.

\subsubsection{Computational Setup} 
\label{sec:computational_setup}

DeepSDMs were trained on a single NVIDIA H100 GPU. Single-species baselines were trained on CPUs, parallelized across species on a compute cluster. Reported training times for single-species models represent the sum of per-species CPU time across all \num{5771} species, while times for multi-species DeepSDMs correspond to wall-clock training time on a single GPU. These numbers are therefore not directly comparable and should be interpreted as indicative of the practical computational cost of each method under its typical computational setup.

\section{Results}
\begin{table}[t]
\caption{\textbf{Models performance comparison on \DATASETNAME}. Overall and group-level AUROC ($\%$) are reported as the mean\,±\,standard deviation over 5 random seeds. Training times are given in minutes (see Section \ref{sec:computational_setup} for details). For the corresponding AUPRG performance, see Appendix \ref{sec:appendix_auprg}.}
\centering
\renewcommand{\arraystretch}{1.25}
\setlength{\tabcolsep}{6pt}
\small
\begin{tabular}{@{}l c cc cc c@{}}
\toprule
& \multicolumn{5}{c}{\textbf{AUROC (\%) $\uparrow$}} & \\
\cmidrule(lr){2-6}
& & \multicolumn{2}{c}{\textbf{Sparse}} 
& \multicolumn{2}{c}{\textbf{Dense}} & \\
\cmidrule(lr){3-4} \cmidrule(lr){5-6}
\textbf{Model} 
& \textbf{Overall} 
& \textbf{Infrequent} & \textbf{Frequent} 
& \textbf{Infrequent} & \textbf{Frequent}
& \textbf{Time (min)} \\
\midrule
\multicolumn{7}{@{}l}{\textbf{Single-species SDMs}} \\[2pt]
\quad GLM           & 84.0\pms{0.0}            & 84.5 & 81.9 & 86.3 & 83.2 & 1667\pms{25}     \\
\quad BRT           & 85.2\pms{0.0}            & 85.5 & 83.5 & 87.2 & 84.6 & 1152\pms{29}     \\
\quad GAM           & 83.8\pms{0.1}             & 83.9 & 82.0 & 85.9 & 83.4 & 2073\pms{49}     \\
\quad MaxEnt        & 84.5\pms{0.0}             & 84.7 & 82.8 & 86.6 & 83.8 & 12383\pms{242} \\
\quad RF & \textbf{85.9}\pms{0.0} & \textbf{86.6} & \textbf{84.3} & \underline{87.7} & \textbf{85.0} & 2610\pms{35}    \\
\midrule
\multicolumn{7}{@{}l}{\textbf{Multi-species DeepSDMs}} \\[2pt]
\quad Baseline   & 83.3\pms{0.1}          & 82.3 & 80.5 & 86.4 & 83.8 & 183\pms{6} \\
\quad + Aggregation & 84.0\pms{0.0}          & 84.7 & 81.7 & 86.3 & 83.3 & 64\pms{5}  \\
\quad + Subsampling      & 84.2\pms{0.1}          & 84.9 & 81.8 & 86.8 & 83.5 & 26\pms{2}   \\
\quad + Architecture     & 84.5\pms{0.0}          & 85.0 & 82.1 & 87.1 & 83.9 & 19\pms{2}   \\
\quad + Loss             & \underline{85.8}\pms{0.0} & \underline{86.5} & \underline{83.8} & \textbf{87.9} & \underline{84.9} & 17\pms{1}   \\
\bottomrule
\end{tabular}
\\
\vspace{5pt}
\raggedright
\label{tab:main_results}
\end{table}

\subsection{Overall Model Performance}

We first compare the aggregated performance of the different modeling approaches across all species (Table~\ref{tab:main_results}). Random Forest achieves the highest performance among the single-species approaches, followed by BRT and MaxEnt. The GAM and GLM exhibit the lowest predictive performance, likely reflecting their limited ability to capture complex non-linear relationships.

Comparing these results to those of multi-species DeepSDMs, we observe that the DeepSDM baseline underperforms compared to single-species models. However, successive design choices significantly improve results. Each design choice contributes positively to overall performance, with data aggregation and the loss function producing the largest gains. Finally, applying all proposed design choices results in a performance comparable to that of the Random Forest. 

In addition to improving predictive performance, the design choices considerably improve computational efficiency. Training a single multi-species model, rather than one model per species, avoids fitting thousands of separate models. The data-related design choices further reduce the per-epoch computation time, while model and training refinements lead to faster convergence; together, these changes reduce computation by one order of magnitude relative to the baseline DeepSDM. Training times are not directly comparable across model families, however, as the single-species and multi-species models are run on different hardware (Section~\ref{sec:computational_setup}).

\subsection{Model Performance grouped by Species Data Properties}

\begin{figure}
    \centering
    \begin{subfigure}[b]{0.99\textwidth}
        \centering
        \includegraphics[width=\textwidth]{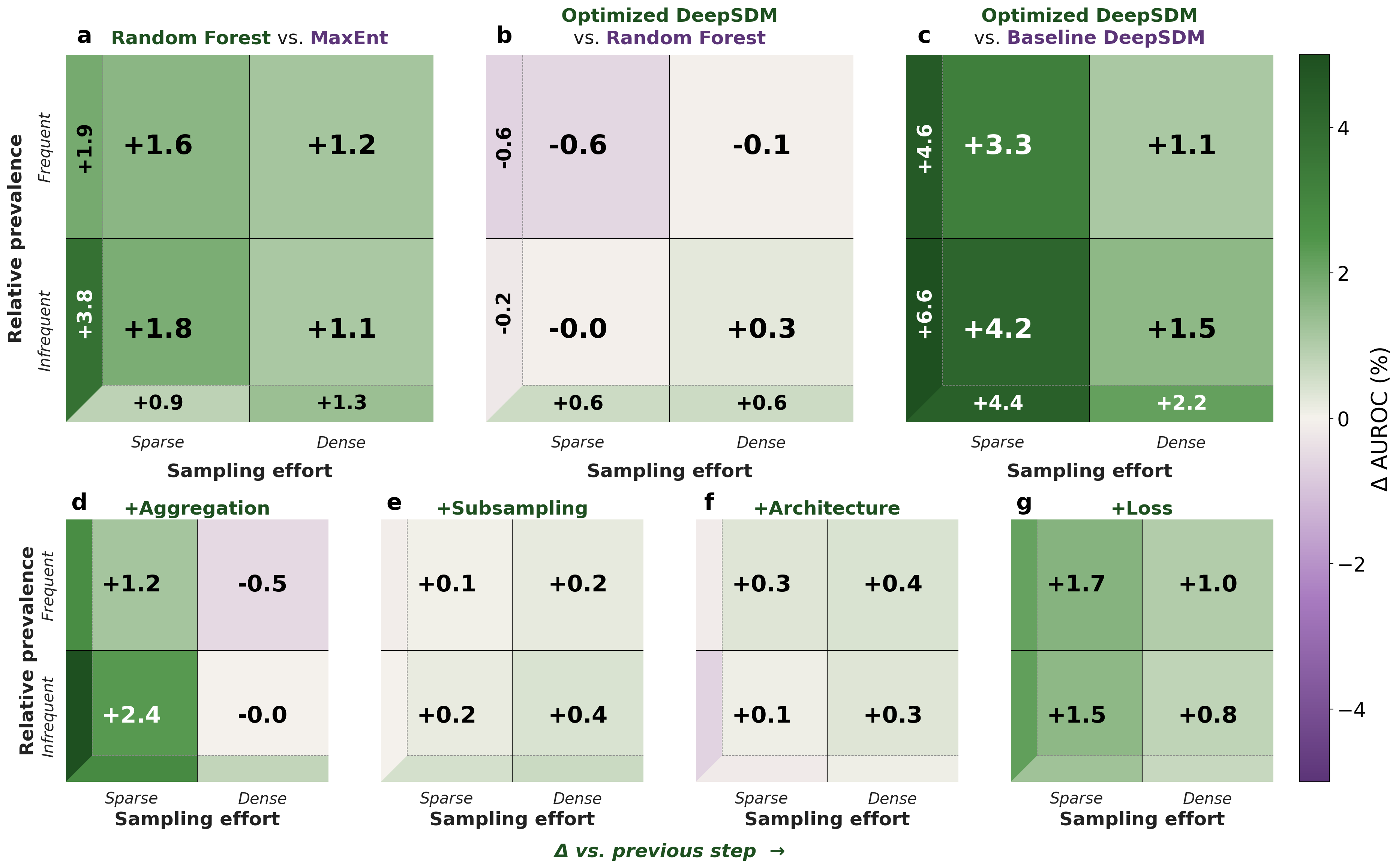}
    \end{subfigure}
    \caption{\textbf{Performance differences ($\Delta$AUROC, \%) across species groups in the data-property space}, with sampling effort on the horizontal axis and relative prevalence on the vertical axis. The thin dashed L-strips along the bottom and left edges isolate the tails of the distribution (bottom: the 10\% of species with the most sparsely sampled ranges; left: the 10\% of species most infrequently observed). \textbf{(a--c)} Pairwise model comparisons: (a) Random Forest -- MaxEnt, (b) Optimized DeepSDM -- RF, (c) Optimized DeepSDM -- Baseline DeepSDM. Positive values (green) indicate that the model named first in each panel title outperforms the one named second; negative values (purple) indicate the opposite. \textbf{(d--g)} Incremental effect of each design choice with respect to the previous configuration considered for the DeepSDM setup: (d) aggregation, (e) subsampling, (f) architecture tuning, (g) loss function. Positive values (green) indicate that the added design choice outperforms the previous configuration, while negative values (purple) indicate the reverse.}
    \label{fig:quadrants}
\end{figure}

We use the species data properties defined in Section~\ref{sec:evaluation} to assess how model choices affect each data-deficiency regime. 
The performance difference between Random Forest and MaxEnt is consistent across species groups (Figure~\ref{fig:quadrants}a), with the largest gains for species with sparse sampling effort. Comparing the optimized DeepSDM with the Random Forest reveals a more nuanced picture (Figure~\ref{fig:quadrants}b): the DeepSDM performs better for infrequently recorded species, whereas the Random Forest is stronger across sparsely sampled regions, most clearly for sparse but frequently recorded species.
The optimized DeepSDM, which incorporates all proposed design choices, improves performance over the DeepSDM baseline across all species groups (Table~\ref{tab:main_results}). However, the magnitude of these improvements varies across groups. In particular, gains are larger for species with sparse sampling effort, highlighting the effectiveness of the design choices in mitigating the impact of sampling bias. These improvements are even more pronounced for species that are both sparsely sampled and infrequent (Figure~\ref{fig:quadrants}c). The largest boost in performance ($+6.7\%$) is found for very infrequently sampled species in extremely sparsely sampled areas (the lowest decile of relative prevalence and sampling effort).

\subsection{Impacts of the Design Choices on Model Performance}

\begin{table*}[ht]                                                   \caption{\textbf{Sensitivity analyses of the design choices.} Average AUROC ($\%$) across all species for different (a) subsampling thresholds, (b) model architectures, and (c) loss configurations. Here, $d$ denotes the hidden dimension, $B$ the number of blocks, and the loss weighting can be based on the frequency of species presences and/or fine-tuned via its hyperparameters.}
  \centering
  \renewcommand{\arraystretch}{1.2}
  \small
    \begin{tabular*}{\textwidth}{@{\extracolsep{\fill}} lc lllc llc @{}}       
  \multicolumn{2}{c}{(a) \textbf{Subsampling}} & \multicolumn{4}{c}{(b) \textbf{Architecture}} & \multicolumn{3}{c}{(c) \textbf{Loss}} \\
  \cmidrule(r){1-2} \cmidrule(lr){3-6} \cmidrule(l){7-9} Threshold $N$ & AUROC & Model & $d$ & $B$ & AUROC & Pseudo-absence & Weighting &  AUROC \\
  \cmidrule(r){1-2} \cmidrule(lr){3-6} \cmidrule(l){7-9}
  None (full) & 84.0 & MLP & 512 & 4 & 84.3 & Both & None  & 84.5 \\ %
  \num{100000} & 84.0 & ResNet & 512 & 4 & 84.4 & Random & None  & 80.1 \\ 
  \num{10000} & 84.1 & ResNet & 512 & 8 & 84.5 & Target-group & None  & 85.5 \\
  \num{1000} & \textbf{84.3} & ResNet & 1024 & 8 & 84.5 & Target-group & Tuned  & 85.7 \\ %
  \num{100} & 83.7 & Transf. & 512 & 4 & 84.5 & Target-group & Freq. & 85.7 \\ %
  \num{10} & 80.8 & Transf. & 1024 & 8 & \textbf{84.8} & Target-group & Freq.+Tuned & \textbf{85.8} \\ %
  \cmidrule(r){1-2} \cmidrule(lr){3-6} \cmidrule(l){7-9}
  \end{tabular*}
  \label{tab:ablation}

\end{table*}

We examine how each choice affects different species groups (Figure \ref{fig:quadrants}d--g) and analyze hyperparameter sensitivity of the design choices (Table \ref{tab:ablation}). Sensitivity analyses are conducted sequentially: once the best configuration for a given design choice is identified, it is carried over into all subsequent analyses. 

\paragraph{Aggregation.}
Data aggregation is the most effective data-related design choice, while strongly reducing training time (Table \ref{tab:main_results}). Its effect is uneven across species groups: species with low sampling effort benefit from aggregation, with the largest improvements in AUROC observed for infrequent species in sparsely sampled areas (Figure \ref{fig:quadrants}d). In contrast, performance decreases for species in densely sampled areas when using aggregated data. No sensitivity analysis was performed for this design choice because we considered data aggregation only at the coarsest resolution of the predictors (1~km).

\paragraph{Subsampling.}
Limiting the number of occurrences per species reduces the influence of heavily sampled taxa on model training, while limiting the number of samples lowers computational cost (Table \ref{tab:main_results}). We find that model performance improves marginally, albeit steadily, as we decrease the subsampling threshold down to \num{1000} occurrences per species (Table \ref{tab:ablation}a). Below this threshold, performance declines more sharply. We note that, with a threshold of \num{1000}, $1.6$M samples remain out of $4.1$M in the aggregated dataset. Remarkably, this reduction of over 60\% yields the best overall score, indicating that a relatively small number of well-distributed samples already captures species-environment relationships effectively. Across species groups, subsampling provides consistent improvements, irrespective of sampling effort or relative prevalence (Figure \ref{fig:quadrants}e).

\paragraph{Architecture.} 
We compare our Baseline MLP to residual networks and FT-Transformers of varying depth and width, defined by the number of blocks and the hidden dimension, respectively. Switching from an MLP to a ResNet results in a minimal improvement, and also increasing the model size yields only a marginal improvement (+0.2\%, Table \ref{tab:ablation}b), suggesting that for tabular environmental predictors, model architecture and capacity are not the primary bottlenecks. The Transformer slightly outperforms the ResNet (+0.3\%), but requires substantially higher computational demands, on the order of $35\times$ more training time. We therefore use the best ResNet for the main results. Compared to the baseline MLP, this improvement is consistent across species groups, except for species in very sparsely sampled ranges (the lowest decile of sampling effort), where we observe a slight decrease in performance (Figure \ref{fig:quadrants}f).

\paragraph{Loss.}
We examine how different loss configurations affect model performance by varying the type of pseudo-absences and the weighting strategy from \cite{zbinden2024selection}. 
The most influential factor is the type of pseudo-absences (Table \ref{tab:ablation}c). Using target-group background instead of random background leads to a substantial improvement in AUROC ($+5.3\%$), while combining both types yields intermediate performance ($+4.4\%$). The choice of weighting strategy has a more modest effect. Both fine-tuned and frequency-based weighting improve over uniform weighting and achieve comparable performance, with frequency-based weighting providing a practical solution, avoiding additional hyperparameter tuning. We therefore use the frequency-based weighting scheme in the main results and find the greatest gains for species with low sampling effort (Figure \ref{fig:quadrants}g).

\paragraph{} Among all design choices, the loss function has the largest individual impact, followed by data aggregation, subsampling, and architecture (Table~\ref{tab:main_results}). Notably, the most influential factors relate to data preparation and optimization rather than model capacity, suggesting that for SDMs using tabular predictors, how the occurrence data is prepared and weighted matters more than the choice of network.
Each design choice also affects species groups differently: aggregation and the loss function primarily benefit undersampled and infrequently recorded species, while subsampling and architecture improve performance more uniformly (Figure \ref{fig:quadrants}d-g). Combined, the four choices yield a cumulative gain of 2.5\% in AUROC over the baseline (Table~\ref{tab:main_results}).

For simplicity, we report only AUROC scores throughout the Results section. The corresponding AUPRG results are presented in Appendix \ref{sec:appendix_auprg} and lead to the same overall conclusions, except that the Random Forest retains a clearer advantage over the optimized DeepSDM with this metric. All approaches achieve broadly similar AUROC scores, although non-negligible differences remain and can translate into markedly different prediction maps. Because AUROC is strongly influenced by factors unrelated to model quality \citep{lobo2008auc, sofaer2019area}, we primarily focus on how approaches differ across species groups rather than on the magnitude of accuracy gaps. We return to these metric limitations in the Discussion (Section~\ref{sec:discussion_metrics}).

\subsection{Impact on Prediction Maps}

\begin{figure}[ht]
    \centering
    \begin{subfigure}[b]{0.99\textwidth}
        \centering
        \includegraphics[width=\textwidth]{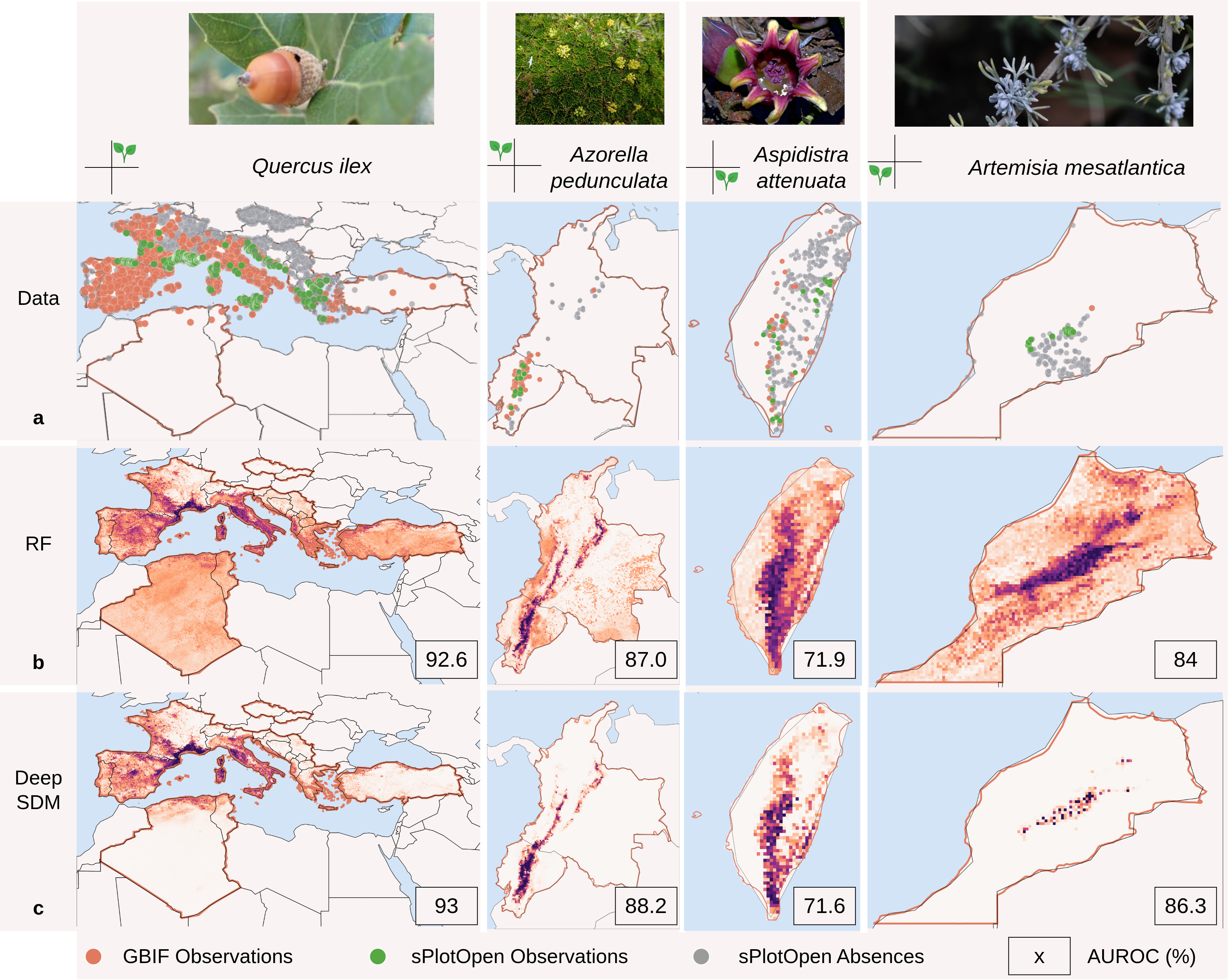}
    \end{subfigure}
    \caption{\textbf{Prediction maps for four species representing the four quadrants of the property space}, comparing Random Forest \textbf{(b)} and the optimized DeepSDM \textbf{(c)}. Predicted probabilities are rescaled for each species and model through min-max normalization; colorbars are therefore not comparable. Training and test data are shown in the top row for reference. Species photos: \textit{Quercus ilex} \copyright~Guillaume Martin, iNaturalist, CC BY-NC 4.0 (observed February 7, 2026), \textit{Azorella pedunculata} \copyright~David Torres, Pl@ntNet, CC BY-NC 4.0 (observed January 18, 2026), \textit{Aspidistra attenuata} \copyright~Jacy Chen, iNaturalist, CC BY 4.0 (observed December 9, 2021), \textit{Artemisia mesatlantica} \copyright~Abdelmonaim Homrani Bakali, teline.fr (Biodiversité végétale du sud-ouest marocain), CC BY-NC 4.0 (observed October 2018).}
    \label{fig:prediction_maps}
\end{figure}

In this section, we illustrate how predictions vary across models for four species, one selected from each quadrant of the property space.
For each species, we compare the prediction maps from the best-performing single-species model (Random Forests) and multi-species model (optimized DeepSDM), along with the corresponding training and evaluation data (Figure \ref{fig:prediction_maps}).

\textit{Quercus ilex} (holly oak or evergreen oak) is a well-sampled species spanning the Mediterranean, with a densely sampled range and high relative prevalence. Both models achieve similar AUROC scores. The predicted distributions are generally in agreement, although the single-species Random Forest finds a stronger signal in areas that have very few occurrences, such as Algeria and Turkey.
\textit{Azorella pedunculata} has low sampling effort in its range but is relatively frequently sampled within the available records. It is reported to be native to Colombia and Ecuador. Here, the single-species Random Forest performs worse compared to the multi-species DeepSDM, with the latter resulting in a more tightly constrained distribution.

\textit{Aspidistra attenuata} (cast-iron-plant or bar-room plant) is reported to be native to Taiwan, which is densely sampled, but occurs only infrequently. The single-species Random Forest produces diffuse suitability across the entire range, resulting in a slightly higher AUROC, whereas the multi-species DeepSDM yields a more structured prediction.
\textit{Artemisia mesatlantica} (blue mugwort) is endemic to Morocco and is twofold data-deficient, as it occurs in a sparsely sampled area and has low relative prevalence, with a single GBIF occurrence record. In this case, the multi-species DeepSDM produces a more localized prediction and achieves a higher AUROC.

These four species should be read as illustrative rather than representative, as performance varies considerably within each quadrant of the property space. 

\subsection{Geographic Distribution of Performance}

\begin{figure}[ht]
    \centering
    \begin{subfigure}[b]{0.99\textwidth}
        \centering
        \includegraphics[width=\textwidth]{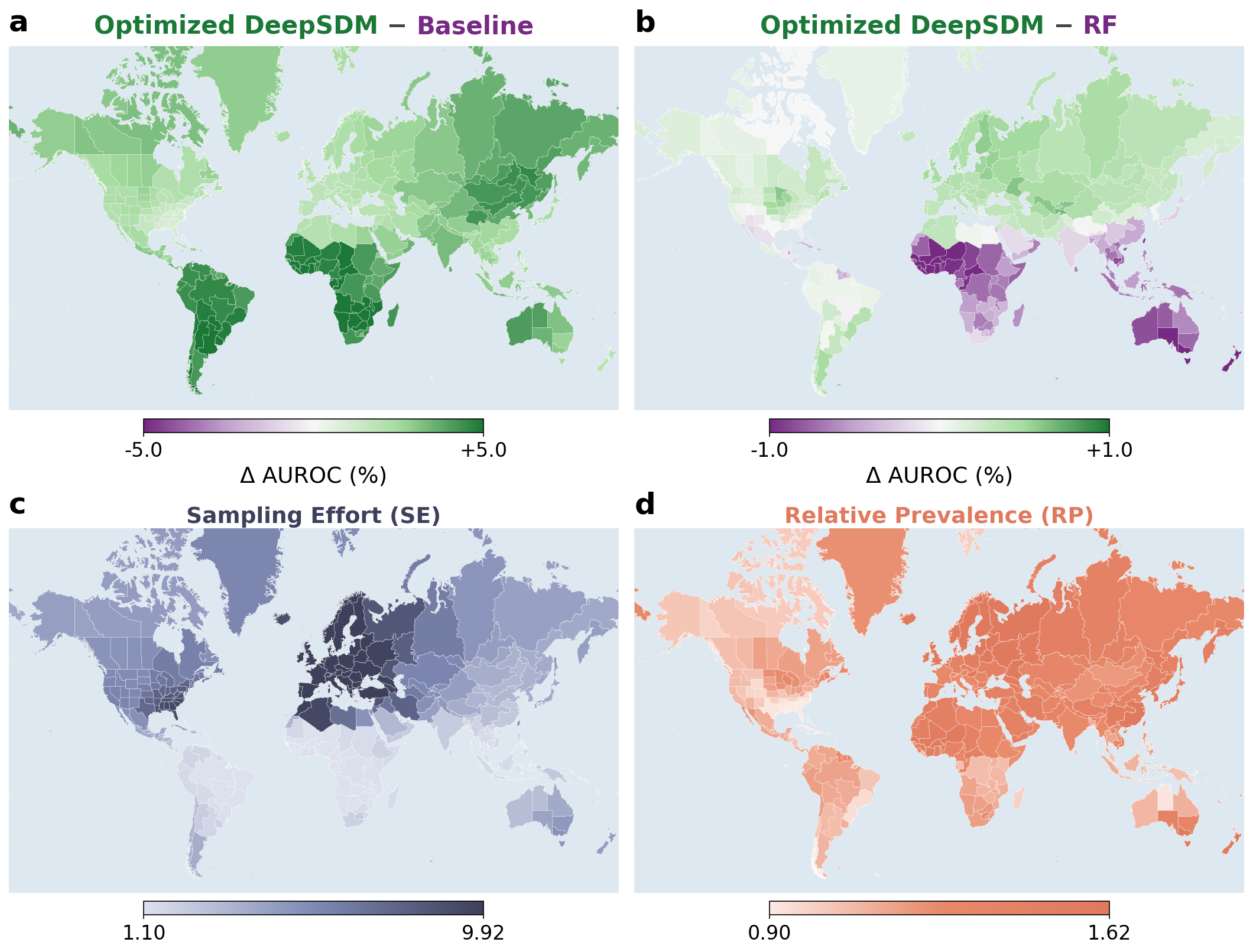}
    \end{subfigure}
    \caption{\textbf{Geographic distribution of model performance and data properties.} Values are obtained by averaging over all species whose ranges overlap each country polygon, or finer administrative units for larger countries. Differences are computed between the optimized DeepSDM and the baseline DeepSDM \textbf{(a)}, and between the optimized DeepSDM and the Random Forest \textbf{(b)}. The two species data properties, sampling effort \textbf{(c)} and relative prevalence \textbf{(d)}, are computed over the same regions. \textit{Note}: the color scale differs between the two performance panels.}
    \label{fig:geographic_delta}
\end{figure}

We observe that performance differences between models depend strongly on the group of species considered. In this section, we further investigate whether these differences are also observed spatially, with some models outperforming others in specific regions. We compute regional performance differences and species data properties by aggregating species-level AUROC changes across all species whose reported native ranges intersect each country polygon, or finer administrative units for larger countries (Figure~\ref{fig:geographic_delta}a,b), and map the two species data properties using the same aggregation scheme (Figure~\ref{fig:geographic_delta}c,d). The resulting maps reveal substantial spatial heterogeneity in model performance gains. Improvements of the optimized DeepSDM over the baseline DeepSDM are widespread, but uneven, with stronger gains concentrated in the regions with the lowest sampling effort, such as South America and Africa (Figure~\ref{fig:geographic_delta}a,c). In contrast, gains over the Random Forest are smaller, more spatially localized, and follow the opposite gradient. The optimized DeepSDM performs better in well-sampled regions, whereas the Random Forest leads across the undersampled tropics, indicating that no single modeling approach uniformly dominates across all geographic contexts. 

The spatial gradient is shaped almost entirely by sampling effort rather than relative prevalence. For both properties, variance lies mainly among species rather than among regions, but this is far more pronounced for relative prevalence. Once aggregated to regions, it retains only $\approx 4\%$ of its variance among regions, compared to roughly $16\%$ for sampling effort (Figure~\ref{fig:geographic_delta}c,d). This is expected, as sampling effort is essentially a geographic property, with some regions more densely surveyed than others, and therefore preserves clear regional structure. Relative prevalence, by contrast, characterizes each species, describing how frequently it is recorded within its range, and is largely averaged out across space. The spatial gradient is therefore shaped almost entirely by sampling effort, and the prevalence-driven differences identified in the property-space analysis (Figure~\ref{fig:quadrants}) do not translate into clear geographic patterns.

\section{Discussion}

\subsection{Model Performance and Comparison}
\label{sec:discussion_metrics}
In this study, we use our benchmark dataset and evaluation framework, \DATASETNAME{}, to compare the performance of several established single-species SDM approaches and multi-species DeepSDMs. Our results illustrate how averaging performance across species can hinder insightful model comparison. Methods with nearly identical average performance may exhibit strong disparities across species, both in predicted distributions and in accuracy, with performance structured by sampling-related properties. Notably, while the average AUROC scores for Random Forests and our optimized DeepSDM are very close, species with low relative prevalence tend to benefit more from the DeepSDM. This supports the hypothesis that joint modeling is particularly beneficial for underrepresented and infrequently recorded species, which can leverage the shared embedding space and co-occurrence signal from better-sampled taxa \citep{zbinden2023exploring, cole2023spatial}. Random Forests, in contrast, remain stronger for species occurring in regions with sparse sampling effort. This may reflect both their robustness as ensembles, a property that has made them consistently strong SDM baselines \citep{valavi2022predictive}, and the independent fitting of each species, which lets the model concentrate fully on its few records rather than diluting them within an objective optimized across thousands of species. Neither approach dominates, and the better choice depends on the species and regions of interest.

Achieving competitive performance with DeepSDMs nonetheless requires careful data and modeling choices, particularly for species with low sampling effort. These design choices---data aggregation, subsampling, model architecture, and loss function---do not uniformly benefit all species, but collectively yield significant improvements, with a total gain of 2.5\% overall. Our results highlight the importance of considering such choices when training DeepSDMs on large-scale, biased multi-species data. 

For instance, when applying data aggregation and subsampling, we were able to obtain equivalent or better model performance using less than 6\% of the samples from the full dataset, thereby requiring much shorter computation time. This highlights that, although DeepSDMs can exploit large amounts of data, smaller, curated datasets that avoid local oversampling can be just as effective while reducing the computational resources. Citizen science datasets are prone to regional and local oversampling \citep{hughes2021sampling, zizka2021sampbias}, which is a form of statistical pseudo-replication. Spatial thinning is an effective sampling bias correction method used to mitigate such oversampling \citep{boria2014spatial, sillero2021want}. The data aggregation and subsampling we propose in this study are analogous to data thinning. The former is equivalent to spatial thinning at the grain of the environmental covariates, while the latter applies random subsampling, mitigating extreme imbalances in the number of occurrences across species. Alternative subsampling strategies in geographic or environmental space can also effectively mitigate data biases \citep{pili2025correcting}, but remain to be assessed in the context of DeepSDMs.

Beyond these data-level corrections, the loss function offers further leverage for addressing sampling bias. The most consequential choice is the type of pseudo-absences: target-group background points yield a large improvement over random background, in line with prior work showing that carefully selected pseudo-absences can help mitigate geographic sampling bias \citep{phillips2009sample, botella2020bias}. The weighting strategy has a smaller but consistent effect. Species-specific weighting of presences and pseudo-absences improves over uniform weighting and addresses the cross-species imbalance in the ratio of background points to observations, which varies by orders of magnitude and is avoided by construction in single-species models \citep{zbinden2024selection}.

The results we present in this study use the AUROC as a performance metric, as it is one of the most commonly used metrics in SDMs and has the advantage of avoiding the need to threshold predictions \citep{li2024area, richardson2024receiver}. However, AUROC is strongly influenced by factors such as prevalence and the presence of easily classifiable samples \citep{lobo2008auc, sofaer2019area}. Therefore, it can be misleading to compare AUROC scores across species, but relative differences in performance between models remain informative and provide a reliable basis for comparison. Nonetheless, additional or alternative metrics may better capture SDM performance in specific contexts and are worth exploring in future work. The corresponding AUPRG results are provided in Appendix~\ref{sec:appendix_auprg}. 

\subsection{Benchmark Data and Evaluation Framework}
By releasing the dataset and code of the evaluation framework behind \DATASETNAME\, we enable future work to assess the performance of other models on the same dataset in the same way, allowing consistent and fair comparison. \DATASETNAME\ is designed to reflect the complexity of biodiversity data in the real-world while remaining broadly usable across modeling approaches.
To ensure comparability of performance metrics across studies, training and evaluation datasets should be retained. In particular, species occurrence data splits for training, validation, and testing must remain identical for comparability, but future work may investigate the effect of other design choices, including alternative predictors. Here, we restrict predictors to tabular environmental variables, i.e., single scalar features per location, to ensure accessibility to both DL and statistical methods. Future work may assess the effect of integrating more complex data sources and modalities. For example, studies have found that incorporating satellite imagery or temporal data in DeepSDMs can improve model performance \citep{teng2023satbird, dollinger2024sat, picek2024geoplant}, and a framework like ours could shed light on how the strength of this effect varies across species with different sampling biases.
Furthermore, because species identities are preserved, the dataset can be linked to rich auxiliary information describing species characteristics and interactions, such as functional traits or food webs. In particular, community-level functional traits from TRY \citep{kattge2020try} are already available through sPlotOpen, enabling extensions such as trait-aware modeling.

The evaluation framework presented in this paper is built upon two properties---sampling effort and relative prevalence---which provide a simple yet effective lens for interpreting performance disparities, as they directly shape the number and distribution of occurrences available for training. We note that the properties considered here were constructed using species ranges derived from reported native countries \citep{powo2025}. Such coarse ranges may overestimate species extents, whereas incomplete checklists may lead to underestimation. Finer-grained expert ranges would address these limitations, but their availability remains limited. Furthermore, these axes represent only a partial view of the underlying biases, and other sources of bias could also be considered. For instance, temporal biases are also present in citizen-science occurrence data, with observation activity often peaking during specific seasons and weekends \citep{courter2013weekend, troudet2017taxonomic, sierra2025divshift}. More broadly, our goal is to encourage a more systematic analysis of how data characteristics influence model performance. Extending \DATASETNAME\ to other species properties could yield additional insights---for instance, by considering the size of the range covered by the occurrence data, or the heterogeneity of its spatial distribution. Future studies may also consider properties computed in the environmental space defined by the predictors \citep{graham2025biodiversity}, rather than spatially as done here.

Our benchmark opens several directions for future work. In particular, the strong disparities of model performance observed across groups of species suggest that adaptive or species-aware modeling strategies could further improve robustness. More broadly, \DATASETNAME\ offers a standardized evaluation framework informed by species data biases and going beyond averaged performance metrics. It provides a foundation for developing novel methodologies and evaluating next-generation biodiversity models that are both scalable and robust to real-world data imperfections. By exposing how sampling bias shapes model performance at scale, this work takes a step toward more reliable, transparent, and ecologically meaningful evaluation of multi-species SDMs.

\section{Conclusion}

Using our Sampling-Aware Global Evaluation benchmark, we evaluate SDMs across thousands of species with highly heterogeneous data characteristics. We release a benchmark dataset representative of large-scale biodiversity datasets that are increasingly available, paired with an evaluation framework designed to consider the severe sampling biases found in these data. \DATASETNAME{} enables comparison of model performance, in particular, investigating how different models or design choices affect predictive performance for species with varying levels of data bias. Evaluating which models can accurately predict species distributions, not only in well-sampled areas for frequently observed species, but also for data-deficient species, is crucial to ensure effective decision-making for conservation efforts.

Our results highlight the need for more informative multi-species evaluation protocols: aggregating metrics can obscure substantial disparities between well-sampled and neglected taxa, providing an incomplete picture of model performance. By explicitly stratifying the evaluation according to species data properties, \DATASETNAME\ enables a more transparent and fine-grained assessment of model behavior. While previous work has emphasized the impact of sampling bias on SDMs, systematic large-scale evaluations remain limited and rarely account for variation in sampling effort and prevalence across species. Our work addresses this gap by combining a global benchmark dataset with a sampling-aware evaluation framework that captures these differences under realistic data conditions. This type of evaluation may also guide model selection based on the biases present in a dataset and the conservation priorities at hand. As biodiversity data continue to grow in scale, so do the biases --- evaluation frameworks such as \DATASETNAME{} are therefore important to ensure that model developments translate into better predictions for all species, not just for the species for which we have the most data.

\section*{Open research statement} The code developed for this study is available at \url{https://github.com/earens/sage-sdm-benchmark}. The processed data and model outputs are available in Zenodo repositories at \url{https://doi.org/10.5281/zenodo.21297133} and \url{https://doi.org/10.5281/zenodo.21335338}, respectively. The latter also contains an archived snapshot of the code repository. Fitted single-species model objects are not distributed due to their size, but are reproducible from the released data and configurations. The GBIF occurrence data used in this study are available at \url{https://doi.org/10.15468/DL.JD7XFA} \citep{gbif2026download}. The sPlotOpen data can be downloaded following the instructions available at \url{https://www.idiv.de/research/projects/splot/splotopen-splot/} \citep{sabatini2021splotopen}. Species reported native countries and corresponding polygons are available from \url{https://powo.science.kew.org/} \citep{powo2025} and \url{https://github.com/tdwg/wgsrpd} \citep{brummitt2001world}, respectively. Bioclimatic, soil, topography, and human footprint predictors are available at \url{https://www.chelsa-climate.org/} \citep{brun2022chelsa}, \url{https://www.soilgrids.org/} \citep{hengl2017soilgrids250m}, \url{https://www.earthenv.org/topography} \citep{amatulli2018suite}, and \url{https://datadryad.org/dataset/doi:10.5061/dryad.052q5} \citep{venter2016global}, respectively.

\section*{Acknowledgments}
NvT, RZ, and DT acknowledge funding from the Swiss National Science Foundation under the deepHSM project (200021\_204057). 
DR acknowledges funding from the Embed2Scale project, which is co-funded by the EU Horizon Europe program under Grant Agreement No 101131841, the Swiss State Secretariat for Education, Research and Innovation (SERI), and UK Research and Innovation (UKRI).
We acknowledge the contributions of the GBIF community, including the institutions, researchers, and observers who collected and shared biodiversity records. We also thank the sPlotOpen initiative and its contributors for making global vegetation plot data accessible for ecological research.

\section*{Author Contributions}
\textbf{EA:} Conceptualization (equal), Data Curation (lead), Formal Analysis (lead), Methodology (equal), Software (lead), Visualization (lead), Writing – Original Draft Preparation (equal), and Review \& Editing. 
\textbf{NvT}: Conceptualization (equal), Methodology (equal), Writing – Original Draft Preparation (equal), and Review \& Editing. 
\textbf{RZ:} Conceptualization (equal), Data Curation (Supporting), Methodology (equal), Software (Supporting), Writing – Original Draft Preparation (equal), and Review \& Editing. 
\textbf{DR:} Conceptualization (supporting), Methodology (supporting), Writing – Review \& Editing. 
\textbf{LD:} Conceptualization (supporting), Writing – Review \& Editing. 
\textbf{CV:} Conceptualization (supporting), Writing – Review \& Editing. 
\textbf{BK:} Conceptualization (supporting), Writing – Review \& Editing. 
\textbf{NEZ:} Writing – Review \& Editing. 
\textbf{LP:} Writing – Review \& Editing. 
\textbf{DT:} Supervision (equal), Writing – Review \& Editing. 
\textbf{JDW:} Supervision (equal), Writing – Review \& Editing.

\section*{Conflict of Interest Statement}
The authors declare no conflicts of interest.

\bibliographystyle{apalike}
\bibliography{references}

\newpage
\appendix
\definecolor{lightgreen}{HTML}{B2DBA9}
\definecolor{lightorange}{HTML}{FDC872}
\definecolor{lightred}{HTML}{F0A59E}

\section*{Appendices}

\noindent These appendices provides additional methodological details and supplementary results supporting the main text.

\begin{itemize}
    \item \textbf{Appendix~\ref{sec:appendix_datasets} -- Additional dataset details}
    \begin{itemize}
        \item \ref{sec:appendix_splotopen}: sPlotOpen
        \item \ref{sec:appendix_gbif}: GBIF
        \item \ref{sec:appendix_data_stats}: Dataset statistics
        \item \ref{sec:appendix_range_maps}: Range maps
        \item \ref{sec:appendix_predictors}: Environmental predictors
    \end{itemize}

    \item \textbf{Appendix~\ref{sec:appendix_evaluation} -- Additional evaluation framework details}
    \begin{itemize}
        \item \ref{sec:appendix_properties}: Species data properties
    \end{itemize}

    \item \textbf{Appendix~\ref{sec:appendix_modeling} -- Additional modeling details}
    \begin{itemize}
        \item \ref{sec:appendix_single_species}: Single-species models
        \item \ref{sec:appendix_multi_species}: Multi-species deep learning models
        \item \ref{sec:appendix_loss}: Loss functions
        \item \ref{sec:appendix_ablation_order}: Order of design choices
    \end{itemize}

    \item \textbf{Appendix~\ref{sec:appendix_results} -- Additional results}
    \begin{itemize}
        \item \ref{sec:appendix_finegrained_performance_heatmap}: Fine-grained performance heatmaps
        \item \ref{sec:appendix_auprg}: AUPRG results
        \item \ref{sec:appendix_plot_area}: Effect of the plot size
    \end{itemize}
\end{itemize}

\section{Additional dataset details}
\label{sec:appendix_datasets}

\subsection{SPlotOpen}
\label{sec:appendix_splotopen}

\paragraph{Initial filtering.} 
We download the sPlotOpen v2 dataset from iDiv \citep{sabatini2021splotopen}, which provides four files: species occurrences with cover values, plot-level metadata (headers), community-weighted trait means, and dataset metadata. We merge the species occurrence table with the plot header table using the shared plot observation identifier. We then apply two filters. First, we retain only plots in which all vascular plants were recorded, as indicated by the \texttt{Plant\_recorded} field taking the value ``All vascular plants'' or ``Not specified''. Only a minority of sPlotOpen plots carry explicit information on the group of plants sampled; for the remaining plots, \cite{sabatini2021splotopen} state that it is ``safe to assume that, unless otherwise specified, plots contain information on all vascular plants.'' Second, we exclude plots with reported location uncertainty exceeding \num{1000}~m, matching the coarsest spatial resolution of our environmental predictors.

\paragraph{Taxonomic harmonization.}
Species names in sPlotOpen follow the original nomenclature of each contributing vegetation dataset, resulting in considerable taxonomic heterogeneity. To ensure consistent species identities across all data sources, we resolve all original species names to accepted names in the World Checklist of Vascular Plants (WCVP; \cite{govaerts2021world}) using the \texttt{rWCVP} R package \citep{brown2023rwcvp}.

The harmonization proceeds in two steps. First, each scientific name string is parsed into a taxon name and an author string, and records are deduplicated by taxon name. Exact matching against the WCVP database is then attempted using both the taxon name and author string for disambiguation. Second, names that fail exact matching are subjected to strict fuzzy matching, where only matches with an edit distance of~1 and a string similarity $\geq 0.95$ are accepted. When a name matches multiple WCVP entries, we resolve the ambiguity by prioritizing accepted names, followed by homotypic synonyms; heterotypic synonyms are discarded to avoid conflating taxonomically distinct entities. For infraspecific taxa (e.g., subspecies and varieties), the accepted parent species is retrieved via the WCVP parent link rather than by string manipulation, ensuring correct species-level assignment even when the parent species name differs from the infraspecific epithet. Names that cannot be resolved are excluded from the dataset.

After harmonization, duplicate records arising from synonymy (i.e., the same species at the same location reported under different original names) are removed.

\paragraph{Native range filtering.}
To restrict the evaluation data to regions where each species is reported as native, we construct native range maps from the WCVP distribution table and TDWG Level~3 regional shapefiles. For each species, we retrieve all TDWG Level~3 regions in which the species is reported as native, excluding records flagged as introduced, extinct, or of doubtful occurrence. The corresponding regional polygons are merged into a single multipolygon geometry per species. We then perform a spatial join, retaining only occurrence records that fall within the native range of their respective species. This step removes records from regions where a species may have been introduced or cultivated, which could otherwise bias the evaluation.

\paragraph{Spatial aggregation to 1km grid.}
To match the resolution of our environmental predictors, we aggregate all records to a 1~km equal-area grid. Coordinates are projected from WGS84 (EPSG:4326) to the equal-area cylindrical projection EPSG:6933, and each record is assigned to its corresponding grid cell. Within each cell, a species is marked as present if it was recorded in at least one contributing survey, even when other surveys in the same cell did not detect it, following \cite{elith2006novel}. Aggregated locations are represented by the centroid coordinates of each grid cell, converted back to WGS84 for storage.

\paragraph{Species selection and empty location recovery.}
After aggregation, we retain only species that occur in at least 30 unique grid cells. This threshold ensures that each species has sufficient occurrences for reliable estimation of evaluation metrics after the validation--test split. Species falling below this threshold are removed from the dataset.

Importantly, grid cells from which all species are removed during filtering (e.g., because they contained only rare or unresolved species) are retained as empty locations in the evaluation dataset.

\paragraph{Validation--test split.}
The curated evaluation dataset is divided into a validation set (20\%) and a test set (80\%). Because species prevalence varies by orders of magnitude, a naive random split could leave rare species with too few occurrences in the test partition. We therefore apply an iterative stratification procedure that processes species from rarest to most common: for each species, a minimum of 20 test-set occurrences is hard-reserved before remaining plots are allocated proportionally.

\subsection{GBIF}
\label{sec:appendix_gbif}

\paragraph{Occurrence retrieval.}
Starting from the list of species retained in the curated sPlotOpen evaluation set, we query the GBIF backbone taxonomy to obtain the corresponding \texttt{speciesKey} identifiers. Species that cannot be matched to a GBIF backbone entry are excluded. We then issue a bulk download request through the GBIF API using the \texttt{pygbif} Python package, retrieving all georeferenced occurrence records associated with the matched species keys. The download was issued on 20 February 2026 and contains \num{326143690} records across \num{11621} constituent datasets \citep{gbif2026download}.

\paragraph{Taxonomic harmonization.}
Because GBIF aggregates records from heterogeneous data providers, a single biological species may appear under multiple scientific names (e.g., different synonyms, author string variants, or outdated nomenclature). We therefore resolve all GBIF \texttt{scientificName} values to accepted species names in WCVP, independently of GBIF's own backbone taxonomy, to ensure taxonomic consistency with the sPlotOpen evaluation data. The harmonization procedure is identical to the one described for sPlotOpen in Appendix~\ref{sec:appendix_splotopen}. Briefly, names are first resolved through exact matching followed by strict fuzzy matching (edit distance~$= 1$, similarity~$\geq 0.95$), with only homotypic synonyms accepted and infraspecific taxa mapped to their parent species using the WCVP parent link. After harmonization, all records corresponding to the same accepted species are merged into a single per-species file. Species that cannot be resolved or that are absent from the sPlotOpen species list are discarded.

\paragraph{Coordinate cleaning.}
We apply a standardized coordinate-cleaning pipeline using the \texttt{CoordinateCleaner} R package \citep{zizka2019coordinatecleaner}. The following filters are applied independently to each species to remove records:
\begin{itemize}
    \item located at country and province centroids, capital cities, and GBIF headquarters;
    \item located at biodiversity institutions (museums, herbaria, zoos);
    \item falling in the ocean (sea test);
    \item located in urban areas;
    \item with identical latitude and longitude values, or coordinates equal to $(0, 0)$;
    \item with exact duplicate coordinates within each species.
\end{itemize}
In addition, we filter records based on coordinate uncertainty, retaining only records with reported \nolinkurl{coordinateUncertaintyInMeters}~$\leq 1000$~m or with missing uncertainty information. We also restrict the dataset to records with a \texttt{basisOfRecord} of \texttt{HUMAN\_OBSERVATION}, \texttt{OBSERVATION}, or \texttt{PRESERVED\_SPECIMEN}, excluding records from machine-generated or material sample sources.

\paragraph{Native range filtering.}
After coordinate cleaning, we further restrict each species' occurrences to its native range, using the same POWO-based range maps described in Appendix~\ref{sec:appendix_range_maps}. Only records falling within the species' native range polygon are retained. This step is applied within the cleaning pipeline via the \texttt{rWCVP} R package, which provides direct access to WCVP distribution data, excluding regions flagged as introduced, extinct, or of doubtful occurrence. Species for which no valid native range can be retrieved, or for which no occurrences remain after range filtering, are logged and excluded from the dataset.

\paragraph{Species alignment.}
After cleaning, we align the GBIF and sPlotOpen species lists by removing from sPlotOpen all species absent from the cleaned GBIF data. Grid cells from which all species are removed during this alignment are preserved as empty locations in the evaluation set, following the same procedure as in Appendix~\ref{sec:appendix_splotopen}.

\paragraph{Data aggregation.}
To reduce spatial autocorrelation arising from clustered sampling and to ensure a consistent spatial grain across species, we construct a dataset variant in which GBIF occurrence records are aggregated to a 1~km equal-area grid (EPSG:6933), following the same procedure as for sPlotOpen. Within each grid cell, species occurrences are aggregated by union: a species is considered present in a cell if at least one of its records falls within that cell. Environmental predictor values are aggregated by averaging all original values within the cell. Aggregated locations are stored as grid cell centroid coordinates in WGS84. 

\begin{figure}[ht]
    \centering
    \begin{subfigure}[b]{0.99\textwidth}
        \centering
        \includegraphics[width=\textwidth]{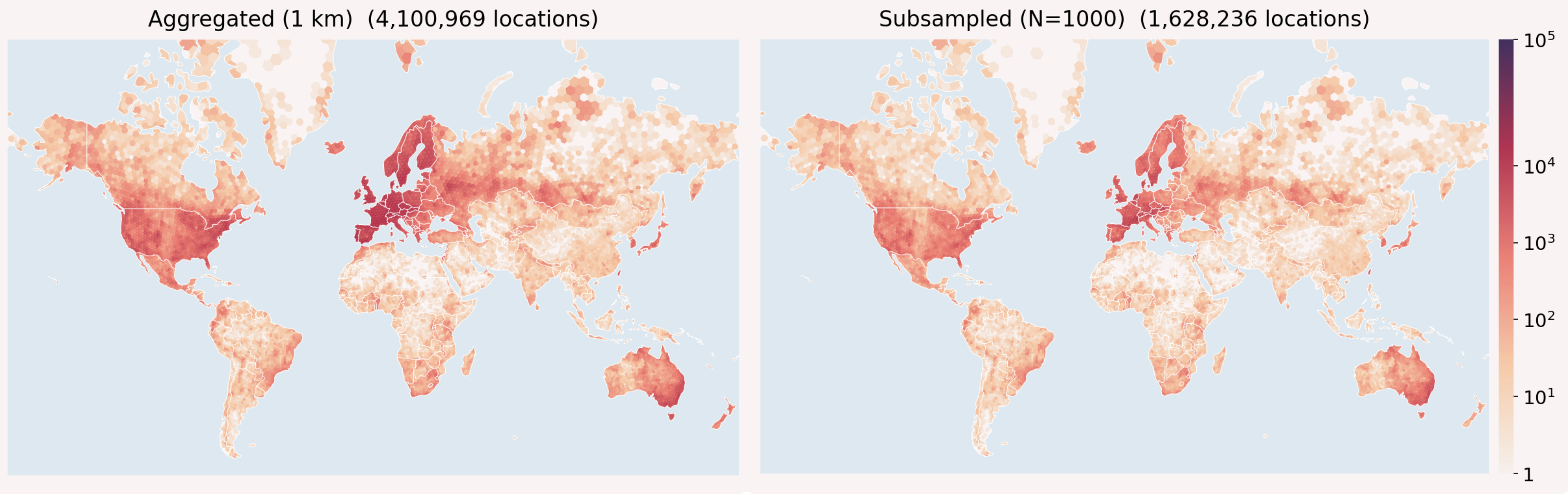}
    \end{subfigure}
    \caption{\textbf{Geographic distribution of occurrence records across the two other GBIF training dataset variants}: aggregated to a 1~km grid (\num{57084622} records; \num{4100969} samples), and subsampled with $N = 1000$ occurrences per species (\num{43071245} records; \num{1627928} samples). Color indicates observation density per H3 hexagonal cell (resolution~3) on a logarithmic scale. All variants retain the same \num{5771} species.}    
    \label{fig:appendix_dataset_maps}   
\end{figure}

This step reduces the GBIF dataset from \num{89768872} across \num{30138089} samples (i.e., distinct coordinates) to \num{57084622} occurrence records across \num{4100969} samples while retaining all \num{5771} species. The geographic distribution of this variant is shown in the left panel of Figure \ref{fig:appendix_dataset_maps}.

\begin{figure}[ht]
    \centering
    \begin{subfigure}[b]{0.99\textwidth}
        \centering
        \includegraphics[width=\textwidth]{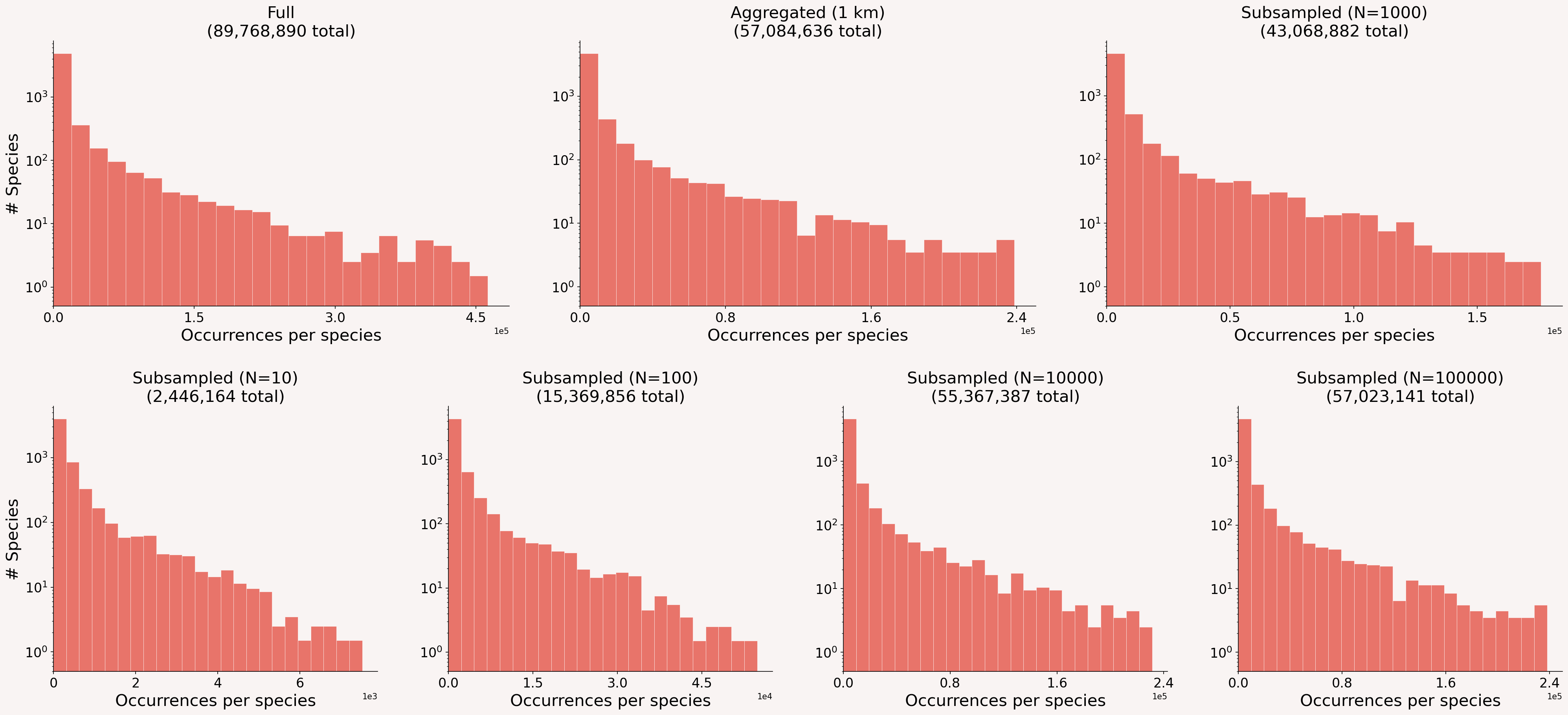}
    \end{subfigure}
    \caption{\textbf{Distribution of per-species occurrence counts across dataset variants.} \textbf{Top row}: full, aggregated (1~km), and subsampled ($N = 1000$). \textbf{Bottom row}: additional subsampling thresholds ($N \in \{10, 100, 10000, 100000\}$). Each panel shows the number of species as a function of occurrence count.}    
    \label{fig:appendix_dataset_hist}   
\end{figure}

\paragraph{Subsampling.}
To further reduce computational cost and mitigate severe imbalances in occurrence counts across species, we construct subsampled dataset variants by limiting the number of occurrence locations per species. For each species, up to $N$ locations are sampled uniformly at random without replacement from the aggregated dataset using a fixed random seed for reproducibility. Because each location records all co-occurring species, a location selected for one species also contributes observations for every other species present at that site, and the same location may be selected for multiple species. As a result, species typically retain more than $N$ occurrence records in the final subsampled dataset. The final occurrence-count distributions across subsampling thresholds are shown in Figure~\ref{fig:appendix_dataset_hist}.

For the main experiments, we use a threshold of $N = 1000$ occurrences per species, reducing the dataset to \num{43071245} occurrence records across \num{1627928} samples while retaining all \num{5771} species (see the geographic distribution in the right panel of Figure \ref{fig:appendix_dataset_maps}). We also evaluate different thresholds ($N \in \{10, 100, 1000, 10000, 100000\}$) in the sensitivity analysis (Table~\ref{tab:ablation}).

\subsection{Dataset Statistics.}
\label{sec:appendix_data_stats}

Table~\ref{tab:dataset_stats} reports detailed statistics for the final dataset, overall and stratified by the four species groups defined in Section~\ref{sec:evaluation}. Both PO occurrences and PA presences are highly concentrated: the dense-frequent quadrant contains roughly a quarter of all species (\num{1422}) yet accounts for about two-thirds of the PO occurrence records (\num{37946692} of \num{57084622} for the aggregated variant), whereas the equally large sparse-infrequent group holds only \num{1927774} records (about 3\%). The PA presences follow the same ordering. The dense-frequent group again contributes the most (\num{433808}; about 47\%) and the sparse-infrequent group the fewest (\num{115671}; about 13\%), but the imbalance is markedly weaker.

  \begin{table}[t]
  \caption{\textbf{Detailed dataset statistics for \DATASETNAME}, overall and stratified by the four species groups defined by the median of sampling effort (sparse/dense) and relative prevalence
  (infrequent/frequent). PO occurrence counts are reported both for the full training set and the 1\,km-aggregated variant used to compute the species properties; PA presences are recorded across
  \num{53336} sPlotOpen plots. Per-species averages and medians are computed over the species in each group.}
  \centering
  \renewcommand{\arraystretch}{1.2}
  \setlength{\tabcolsep}{10pt}
  \small    
  \begin{tabular}{@{}l r rr rr@{}}
  \toprule
  & & \multicolumn{2}{c}{\textbf{Sparse}} & \multicolumn{2}{c}{\textbf{Dense}} \\
  \cmidrule(lr){3-4}\cmidrule(lr){5-6}
  \textbf{Statistic} & \textbf{Overall} & \textbf{Infreq.} & \textbf{Freq.} & \textbf{Infreq.} & \textbf{Freq.} \\
  \midrule
   \# species & \num{5771} & \num{1421} & \num{1464} & \num{1464} & \num{1422} \\
  \midrule
  \multicolumn{6}{@{}l}{\textbf{Presence-only (training)}} \\[2pt]
  \quad \# occurrences (full)   & \num{89768872} & \num{2647373}  & \num{20489219} & \num{5902674}  & \num{60729606} \\
  \quad \# occurrences (1\,km)  & \num{57084622} & \num{1927774}  & \num{13293634} & \num{3916522}  & \num{37946692} \\
  \quad avg.\ / species         & \num{9892}     & \num{1357}     & \num{9080}     & \num{2675}     & \num{26685} \\
  \quad median / species        & \num{1626}     & \num{403}      & \num{1532}     & \num{1306}     & \num{11851} \\
  \midrule
  \multicolumn{6}{@{}l}{\textbf{Presence-absence (evaluation)}} \\[2pt]
  \quad \# presences            & \num{922572}   & \num{115671}   & \num{230931}   & \num{142162}   & \num{433808} \\
  \quad avg.\ / species         & \num{160}      & \num{81}       & \num{158}      & \num{97}       & \num{305} \\
  \quad median / species        & \num{74}       & \num{54}       & \num{78}       & \num{64}       & \num{149} \\

  \bottomrule
  \end{tabular}
  
  \label{tab:dataset_stats}
  \end{table}

\subsection{Range Maps}
\label{sec:appendix_range_maps}

Because our benchmark retains non-anonymized species names, we can integrate external species-level biogeographic information to constrain the spatial domain of each taxon and compute the species data properties used later in the analysis. To achieve complete taxonomic coverage, we rely exclusively on distributional information from POWO \citep{powo2025}.

POWO provides expert-curated presence information at the standardized TDWG Level-3 regional scale. Although these checklist-based products do not represent continuous range boundaries, they offer consistent, global coverage for all vascular plant species in our evaluation list. For each taxon, we aggregate all Level-3 regions in which it is reported and construct a corresponding global range mask. These masks serve two purposes in the benchmark pipeline: (i) restricting spatial prediction and background sampling to continents and regions where the species is known to occur, and (ii) supporting the computation of species data properties.

More fine-grained range polygons are available from the International Union for Conservation of Nature (IUCN), but they only exist for a small subset of vascular plant species. Hence, their coverage is insufficient for a benchmark of this scale and would introduce systematic availability bias. We therefore do not incorporate IUCN shapes into the construction of the benchmark datasets.

\paragraph{Range polygon construction.}
For each species, we retrieve the corresponding accepted name entry from the WCVP names table, restricting to accepted species-rank taxa. We then query the WCVP distribution table for all TDWG Level~3 area codes associated with that species, excluding records flagged as introduced, extinct, or of doubtful occurrence. The corresponding Level~3 regional polygons are retrieved from the TDWG World Geographical Scheme for Recording Plant Distributions shapefile \citep{brummitt2001world} and merged into a single multipolygon geometry per species via a geometric union. The resulting shapefiles are stored on disk and reused across the sPlotOpen and GBIF pipelines.

\paragraph{Spatial indexing for efficient masking.}
To enable efficient range-based filtering during training and evaluation, we precompute two complementary index structures that map between locations and species ranges.

For each occurrence or background location, we determine which TDWG Level~3 region it falls in via a spatial point-in-polygon query against the Level~3 shapefile, and then retrieve all species whose ranges include that region. The result is a compressed index that, for each location, stores the list of species whose native range covers it. During training, this index is used to restrict, for each sample, which species contribute to the loss computation: species whose range does not cover a given location receive an ignore label.

Conversely, for each species, we store the list of all locations falling within its native range. This second index supports efficient species-level operations during evaluation, such as restricting metric computation to locations within the species' range, and is also used for computing the species data properties (sampling effort and relative prevalence) described in the main text.

\subsection{Predictors}
\label{sec:appendix_predictors}

We assemble 52 environmental predictors from four complementary data sources: 19 bioclimatic variables from CHELSA v2.1 \citep{karger2017climatologies}, 8 soil properties from SoilGrids250m \citep{hengl2017soilgrids250m}, 16 topographic variables from the EarthEnv global terrain dataset \citep{amatulli2018suite}, and 9 human footprint variables from the Global Human Footprint dataset \citep{venter2016global}. The complete lists, descriptions, and units of the variables are provided in Tables~\ref{tab:predictors_bioclim}--\ref{tab:predictors_human}.

\begin{table}[ht]
\caption{\textbf{Bioclimatic predictors} derived from CHELSA v2.1 at 1~km spatial resolution \citep{brun2022chelsa}.} %
\centering
\small
\renewcommand{\arraystretch}{1.15}
\begin{tabular}{llp{6cm}l}
\hline
Acronym & Name & Description & Unit \\
\hline
bio1  & Annual mean temperature & Mean annual near-surface air temperature calculated as the average of mean monthly temperatures over the year & $^\circ$C \\
bio2  & Mean diurnal range & Mean diurnal near-surface air temperature range computed as the average of monthly (tasmax $-$ tasmin) & $^\circ$C \\
bio3  & Isothermality & Isothermality: $100 \times (\text{bio2} / \text{bio7})$; compares day--night variability to annual temperature range & unitless \\
bio4  & Temperature seasonality & Standard deviation of mean monthly temperatures & $^\circ$C$/100$ \\
bio5  & Max temp. warmest month & Highest monthly mean of daily maximum temperatures across the year & $^\circ$C \\
bio6  & Min temp. coldest month & Lowest monthly mean of daily minimum temperatures across the year & $^\circ$C \\
bio7  & Temperature annual range & Annual temperature range computed as bio5 $-$ bio6 & $^\circ$C \\
bio8  & Mean temp. wettest quarter & Average monthly mean temperature over the wettest 3-month period of the year & $^\circ$C \\
bio9  & Mean temp. driest quarter & Average monthly mean temperature over the driest 3-month period of the year & $^\circ$C \\
bio10 & Mean temp. warmest quarter & Average monthly mean temperature over the warmest 3-month period of the year & $^\circ$C \\
bio11 & Mean temp. coldest quarter & Average monthly mean temperature over the coldest 3-month period of the year & $^\circ$C \\
bio12 & Annual precipitation & Sum of monthly precipitation totals across the year & kg\,m$^{-2}$\,year$^{-1}$ \\
bio13 & Prec. wettest month & Maximum monthly precipitation total & kg\,m$^{-2}$\,month$^{-1}$ \\
bio14 & Prec. driest month & Minimum monthly precipitation total & kg\,m$^{-2}$\,month$^{-1}$ \\
bio15 & Precipitation seasonality & Coefficient of variation ($100 \times$ SD / mean) of monthly precipitation totals & unitless \\
bio16 & Prec. wettest quarter & Average monthly precipitation during the wettest 3-month period of the year & kg\,m$^{-2}$\,month$^{-1}$ \\
bio17 & Prec. driest quarter & Average monthly precipitation during the driest 3-month period of the year & kg\,m$^{-2}$\,month$^{-1}$ \\
bio18 & Prec. warmest quarter & Average monthly precipitation during the warmest 3-month period of the year & kg\,m$^{-2}$\,month$^{-1}$ \\
bio19 & Prec. coldest quarter & Average monthly precipitation during the coldest 3-month period of the year & kg\,m$^{-2}$\,month$^{-1}$ \\
\hline
\end{tabular}
\label{tab:predictors_bioclim}
\end{table}

\begin{table}[ht]
\caption{\textbf{Soil property predictors} derived from SoilGrids at 250~m spatial resolution \citep{hengl2017soilgrids250m}.}
\centering
\small
\renewcommand{\arraystretch}{1.15}
\begin{tabular}{llp{7cm}l}
\hline
Acronym & Name & Description & Unit \\
\hline
ORCDRC & Soil organic carbon &
Organic carbon content of the fine earth fraction &
g\,kg$^{-1}$ \\
PHIHOX & Soil pH (H$_2$O) &
Soil pH measured in water (stored as pH $\times 10$) &
pH $\times 10$ \\
SNDPPT & Sand content &
Sand fraction of the fine earth fraction &
w\% \\
SLTPPT & Silt content &
Silt fraction of the fine earth fraction &
w\% \\
CLYPPT & Clay content &
Clay fraction of the fine earth fraction &
w\% \\
BLDFIE & Bulk density &
Bulk density of the fine earth fraction &
kg\,m$^{-3}$ \\
CECSOL & Cation exchange capacity &
Cation exchange capacity of the fine earth fraction &
cmol$_\mathrm{c}$\,kg$^{-1}$ \\
BDTICM & Depth to bedrock &
Depth from the soil surface to bedrock &
cm \\
\hline
\end{tabular}
\label{tab:predictors_soil}
\end{table}

\begin{table}[ht]
\caption{\textbf{Topographic predictors} derived from global terrain products at 1~km spatial resolution \citep{amatulli2018suite}.}
\centering
\small
\renewcommand{\arraystretch}{1.15}
\begin{tabular}{llp{6cm}l}
\hline
Acronym & Name & Description & Unit \\
\hline
elevation & Elevation & Terrain elevation & m \\
roughness & Roughness & Largest local elevation difference & m \\
tri & Terrain Ruggedness Index & Mean local elevation difference & m \\
tpi & Topographic Position Index & Elevation relative to local neighbourhood & m \\
vrm & Vector Ruggedness Measure & Local dispersion of surface-normal vectors & unitless \\
aspectcosine & Aspect cosine & Cosine-transformed aspect & unitless \\
aspectsine & Aspect sine & Sine-transformed aspect & unitless \\
slope & Slope & Terrain slope & degrees \\
eastness & Eastness & East--west aspect-slope component & unitless \\
northness & Northness & North--south aspect-slope component & unitless \\
pcurv & Profile curvature & Curvature along the slope direction & rad\,m$^{-1}$ \\
tcurv & Tangential curvature & Curvature perpendicular to the slope direction & rad\,m$^{-1}$ \\
dx & First derivative, E--W & First-order partial derivative in E--W direction & unitless \\
dy & First derivative, N--S & First-order partial derivative in N--S direction & unitless \\
dxx & Second derivative, E--W & Second-order partial derivative in E--W direction & m$^{-1}$ \\
dyy & Second derivative, N--S & Second-order partial derivative in N--S direction & m$^{-1}$ \\
\hline
\end{tabular}
\label{tab:predictors_topo}
\end{table}

\begin{table}[ht]
\caption{\textbf{Human footprint predictors} describing anthropogenic pressures. Variables are derived following the Human Footprint methodology \citep{venter2016global} at 1~km spatial resolution.}
\centering
\small
\renewcommand{\arraystretch}{1.15}
\begin{tabular}{llp{7cm}l}
\hline
Acronym & Name & Pressure Description & Unit \\
\hline
Built2009 & Built environments &
Urban and built-up areas &
score \\
Popdensity2010 & Population density &
Human population density &
score \\
Lights2009 & Night-time lights &
Night-time light emissions &
score \\
Croplands2005 & Croplands &
Intensive cropland agriculture &
score \\
Pasture2009 & Pasturelands &
Grazing and pasture lands &
score \\
Roads & Major roadways &
Roads and associated access &
score \\
Railways & Railways &
Railway infrastructure &
score \\
Navwater2009 & Navigable waterways &
Access via navigable waterways &
score \\
HFP2009 & Human Footprint &
Cumulative Human Footprint &
score \\
\hline
\end{tabular}
\label{tab:predictors_human}
\end{table}

\paragraph{Missing-value imputation.}
Locations falling on raster nodata pixels (e.g., coastal points or small islands) are imputed using nearest-neighbor interpolation. We partition the nodata points into spatial tiles of $512 \times 512$ pixels, read each tile with an additional 50-pixel padding buffer, and apply the Euclidean distance transform to identify the nearest valid pixel. Only replacements within a 50-pixel radius are accepted; points with no valid neighbor within this radius are left as NaN and excluded from training. All extracted predictor values are stored as 32-bit floating-point arrays in HDF5 format.

\section{Additional evaluation framework details}
\label{sec:appendix_evaluation}

\subsection{Species Data Properties}
\label{sec:appendix_properties}
\paragraph{Sampling effort and relative prevalence.}
Both properties are computed on the 1~km aggregated grid described in Appendix~\ref{sec:appendix_splotopen}. The number of grid cells within each species' native range is obtained by rasterizing the range polygon onto the same grid. For sampling effort, visited cells are determined using the species-indexed range masks (Section~\ref{sec:appendix_range_maps}): a cell is considered visited if it contains at least one occurrence record from any species in the aggregated training data. For relative prevalence, we count the number of cells in which the focal species was observed, again within its native range.

\begin{figure}[ht]
    \centering
    \begin{subfigure}[b]{0.99\textwidth}
        \centering
        \includegraphics[width=\textwidth]{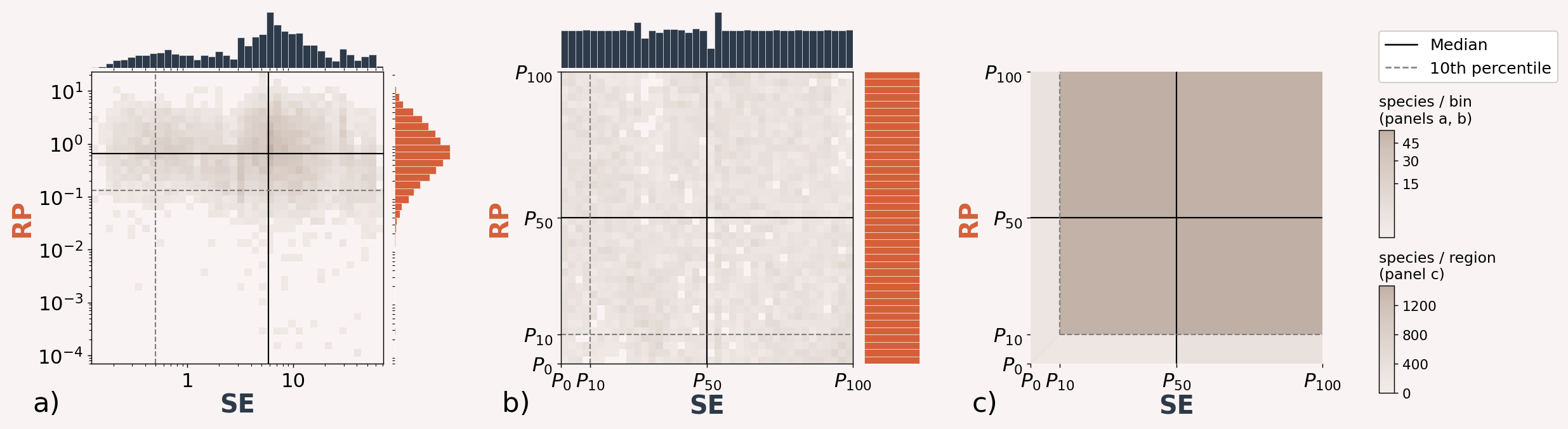}
    \end{subfigure}
    \caption{\textbf{Distribution of species in the two-dimensional property space}, defined by sampling effort (SE) and relative prevalence (RP). \textbf{(a)} Joint distribution shown as a 2D hexbin plot on the original scale, with marginal histograms along each axis. \textbf{(b)} Same distribution shown in percentile-rank space, with percentile thresholds ($P_{10}$, $P_{50}$, $P_{100}$) indicated. \textbf{(c)} The resulting species grouping scheme: four quadrants defined by the median of each property and the lowest-decile groups.}    
     \label{fig:properties_stats}
\end{figure}

Figure~\ref{fig:properties_stats} shows the joint and marginal distributions of both properties across all species, along with the quadrant-based grouping scheme described in Section~\ref{sec:evaluation}.

\paragraph{Independence of the two properties.}
Because sampling effort and relative prevalence share a common term ($|V_s|$, the number of visited cells), one might expect them to be correlated. Empirically, however, they are effectively independent: across all \num{5771} species, the rank (Spearman) correlation is negligible ($\rho = -0.08$), so the two properties share less than $1\%$ of their rank variance. Accordingly, species span nearly the entire percentile-rank space (Figure~\ref{fig:properties_stats}b), confirming that the two properties capture complementary axes of data availability.

\section{Additional modeling details}
\label{sec:appendix_modeling}
\subsection{Single-species models}
\label{sec:appendix_single_species}
We provide implementation details for each single-species baseline. All models are trained independently per species on the subsampled dataset (threshold \num{10000}), with a hard cap of \num{10000} presences per species. Predictors are standardized. Each model receives up to \num{10000} target-group background points drawn from occurrences of other species within the species' range \citep{phillips2009sample}. Because presences are far outnumbered by background, GLM, GAM, and BRT reweight the background contribution to match the number of presences \citep{barbet2012selecting}. Random Forests instead draw a class-balanced bootstrap within each tree 
(down-sampled RF; \citealp{valavi2021modelling}). MaxEnt relies on \texttt{maxnet}'s built-in regularization.

\paragraph{Hyperparameter Selection.} A full per-species grid search over \num{5771} species is impractical, so for each model we sweep a single key hyperparameter over a small grid (Table~\ref{tab:standard_sdm_hyperparameters}) at one seed and, for each species, retain the value with the highest validation AUROC. Holding this per-species choice fixed, we retrain over five seeds and report the mean test performance. MaxEnt is not tuned.

  \begin{table*}[t]
  \caption{\textbf{Hyperparameter grids tested for each standard SDM}, along with the resulting
  overall test AUROC (\%; single reference run, seed 0).}
  \centering
  \label{tab:std_grids}
  \small
  \begin{subtable}[t]{0.24\textwidth}
  \centering\caption{GLM}
  \begin{tabular}{@{}rc@{}}\toprule
  $C$ & AUC \\\midrule
  0.001 & 63.7 \\
  0.01  & 77.3 \\
  0.1   & 82.7 \\
  1     & \textbf{83.5} \\
  10    & 83.5 \\\bottomrule
  \end{tabular}
  \end{subtable}\hfill
  \begin{subtable}[t]{0.24\textwidth}
  \centering\caption{GAM}
  \begin{tabular}{@{}rc@{}}\toprule
  $\lambda$ & AUC \\\midrule
  0.1 & 80.5 \\
  1   & \textbf{84.0} \\
  10  & 83.7 \\
  100 & 82.9 \\\bottomrule
  \end{tabular}
  \end{subtable}\hfill
  \begin{subtable}[t]{0.24\textwidth}
  \centering\caption{BRT}
  \begin{tabular}{@{}rc@{}}\toprule
  tc & AUC \\\midrule
  1 & 84.4 \\
  3 & 85.0 \\
  5 & 85.1 \\
  7 & \textbf{85.1} \\\bottomrule
  \end{tabular}
  \end{subtable}\hfill
  \begin{subtable}[t]{0.24\textwidth}
  \centering\caption{RF}
  \begin{tabular}{@{}rc@{}}\toprule
  mtry & AUC \\\midrule
  $\sqrt{p}$ (7) & \textbf{85.9} \\
  $p/3$ (17)     & 85.7 \\
  $p/2$ (26)     & 85.6 \\\bottomrule
  \end{tabular}
  \end{subtable}\hfill
  \label{tab:standard_sdm_hyperparameters}
  \end{table*}

\paragraph{MaxEnt.}
We use the \texttt{maxnet} R package \citep{phillips2006maximum}. Feature classes are selected adaptively based on the number of presences: linear only ($<$10), linear + quadratic ($<$15), linear + quadratic + product ($<$100), or all classes including hinge and threshold ($\geq$100). We keep \texttt{maxnet}'s defaults: regularization multiplier $\beta=1$ and cloglog output, with clamping at prediction, and do not tune MaxEnt further. We exclude the 11 species with too few data points to train MaxEnt.

\paragraph{Generalized Linear Model (GLM).} We evaluate an L1-regularized (lasso) logistic regression on linear and per-variable quadratic terms using (\texttt{scikit-learn} 1.9.0, \texttt{LogisticRegression}; \citealp{scikit-learn}). We tune the inverse regularization strength $C$ (larger $C$ = weaker penalty) over $\{0.001,0.01,0.1,1,10\}$. Heavy regularization underfits severely ($C{=}0.001$: 63.7) and performance saturates near $C{=}1$ (83.5). 

\paragraph{Generalized Additive Model (GAM).}
Following \cite{valavi2022predictive}, we run a logistic GAM (\texttt{pyGAM} 0.12.0, \texttt{LogisticGAM}; \citealp{daniel_serven_2018_1208723}) with one additive cubic P-spline per predictor (10 basis functions each) and no interaction terms. As for the GLM, we tune the smoothing penalty $\lambda$ (larger = smoother) over $\{0.1,1,10,100\}$. Light smoothing both underperforms and triggers non-convergence for some species (\num{1547}/\num{5771} at $\lambda{=}0.1$). Per-species selection therefore favors more stable settings. Species that fail to converge at all smoothing levels are excluded from the GAM results (30 species).

\paragraph{Boosted Regression Trees (BRT).} For the Gradient Boosting Trees, we run \texttt{LightGBM} 4.7.0 \citep{lightgbm}, configured to emulate the boosted regression trees. We tune the tree complexity, i.e., the number of splits per tree over $\mathrm{tc}\in\{1,3,5,7\}$. Performance increases monotonically with diminishing returns.

\paragraph{Random Forest (RF).} Finally, we train a down-sampled Random Forest (\texttt{imbalanced-learn} 0.14.2, \texttt{BalancedRandomForestClassifier}; \citealp{imbalance}) as suggested by \cite{valavi2021modelling}: \num{1000} fully-grown trees, each fit on a class-balanced bootstrap from the \num{10000}-point background pool, so the ensemble integrates over all background while each tree sees balanced classes. We tune the number of features per split ($\mathit{mtry}$) over $\{7,17,26\}$ corresponding to $\{\sqrt{p},p/3,p/2\}$, where $p=52$ is the number of predictors. Performance is only weakly affected by $\mathit{mtry}$ (85.6--85.9 AUROC), with $\sqrt{p}$ yielding the best results.

\subsection{Multi-species deep learning models}
\label{sec:appendix_multi_species}

All multi-species models are trained with the AdamW optimizer \citep{loshchilov2017decoupled} and the \texttt{ReduceLROnPlateau} learning-rate scheduler from PyTorch, which halves the learning rate when the validation AUROC does not improve for 5 consecutive epochs. Early stopping selects the checkpoint (i.e., the model parameters saved at a particular training epoch) with the highest validation AUROC.

\paragraph{MLP (baseline).}
We evaluate a four-layer multi-layer perceptron with a hidden dimension of 512, trained with a learning rate of $10^{-4}$ and weight decay of $10^{-4}$.

\paragraph{ResNet.}
We adopt the tabular ResNet from \citet{gorishniy2021revisiting} with dropout $0.1$, trained with a learning rate of $10^{-4}$ and weight decay of $10^{-4}$. In the architecture ablation (Table~\ref{tab:ablation}b), we vary the hidden dimension ($d \in \{512, 1024\}$) and the number of residual blocks ($B \in \{4, 8\}$). The best-performing configuration uses $d = 1024$ and $B = 8$.

\paragraph{FT-Transformer.}
We adopt the FT-Transformer architecture proposed by \citet{gorishniy2021revisiting}, using 8 attention heads, a dropout rate of $0.1$, and a learning rate and weight decay of $10^{-4}$. We evaluate a smaller configuration ($d = 512$, $B = 4$) and a larger one ($d = 1024$, $B = 8$), where $d$ denotes the token embedding dimension and $B$ the number of transformer blocks.

\subsection{Loss functions}
\label{sec:appendix_loss}

We describe here in detail the loss functions compared in the corresponding design choice. The loss used in the DeepSDM baseline is defined as follows:

\begin{align} \label{eq:bceloss}
    \mathcal{L}_{\text{baseline}}(\mathbf{y}, \mathbf{\hat{y}}) = - \frac{1}{S} \sum^S_{s=1} \biggl[\mathbbm{1}_{[y_s=1]} \log(\hat{y}_s) + \frac{1}{2}\mathbbm{1}_{[y_s=0]} \log(1 - \hat{y}_s) + \frac{1}{2} \log(1 - \hat{y}_s')\biggr].
\end{align}
Here, $S$ denotes the number of species, $y_s$ is the observed label for species $s$ ($1$ if species $s$ has been observed and $0$ otherwise), $\hat{y}_s$ is the predicted suitability score for species $s$ (ranging from 0 to 1), $\hat{y}_s'$ represents the model's prediction for species $s$ at a random background point location, and $\mathbbm{1}_{[\cdot]}$ is the indicator function, returning $1$ if the condition inside the brackets is true and $0$ otherwise.

This loss consists of three terms: the first accounts for species presences, the second for target-group background points, and the third for random background points. It therefore combines both types of pseudo-absences, following \cite{cole2023spatial}. 

However, this formulation has several limitations, as highlighted by \cite{zbinden2024selection}. First, it does not account for class imbalance: species exhibit highly variable numbers of presences, which also induces variability in the number of associated target-group background points, while random background points are more abundant. Such imbalances are known to degrade performance in machine learning, motivating the use of weighted losses \citep{johnson2019classimbalance}. Second, $\mathcal{L}_{\text{baseline}}$ does not explicitly control the relative importance of the two types of pseudo-absences, despite evidence that target-group background points can be more effective for mitigating sampling bias \citep{phillips2009sample}.

To address these issues, we consider the following loss, based on \cite{zbinden2024selection}:
\begin{equation}
\begin{split}
\mathcal{L}_{\text{optimized}}(\mathbf{y}, \hat{\mathbf{y}})= - \frac{1}{S} \sum_{s=1}^{S} \Biggl[
\mathbbm{1}_{[y_s = 1]} \,\lambda_1\, w_{\mathrm{p}(s)} \log(\hat{y}_s)
&+ \mathbbm{1}_{[y_s = 0]} \,\lambda_2\, w_{\mathrm{tgb}(s)} \log(1 - \hat{y}_s) \\
&+ (1 - \lambda_2)\, w_{\mathrm{rb}(s)} \log(1 - \hat{y}'_s)
\Biggr].
\end{split}
\end{equation}
This formulation introduces \textit{species-specific weights} $w_{\mathrm{p}(s)}$, $w_{\mathrm{tgb}(s)}$, and $w_{\mathrm{rb}(s)}$ to balance contributions across species and between different types of observations. These weights are computed based on the frequency of species observations. The parameter $\lambda_1$ controls the relative importance of presence observations, while $\lambda_2 \in [0,1]$ governs the trade-off between target-group and random background points.

Because samples outside species ranges may be masked during training, we adapt the weighting scheme to reflect only observations within each species’ range. In \cite{zbinden2024selection}, it is assumed that
\(
n_{\mathrm{p}(s)} + n_{\mathrm{tgb}(s)} = n_s = n
\)
where $n_{\mathrm{p}(s)}$ and $n_{\mathrm{tgb}(s)}$ denote the number of presences and target-group background points for species $s$, respectively, and $n$ is the total number of visited sites. Under this assumption, the weights are defined as
\[
w_{\mathrm{p}(s)} = \frac{n_s}{n_{\mathrm{p}(s)}}, \qquad
w_{\mathrm{tgb}(s)} = \frac{n_s}{n_{\mathrm{tgb}(s)}}, \qquad
w_{\mathrm{rb}(s)} = 1 .
\]

In our setting, however, this assumption does not hold because we restrict observations to sites within each species’ range. As a result, $n_s$ varies across species, and the number of random background points $n_{\mathrm{rb}(s)}$ per epoch is also variable, since not all sampled points fall within the species range.

To account for this, we redefine the weights as
\[
w_{\mathrm{p}(s)} = \frac{n}{n_{\mathrm{p}(s)}}, \qquad
w_{\mathrm{tgb}(s)} = \frac{n}{n_{\mathrm{tgb}(s)}}, \qquad
w_{\mathrm{rb}(s)} = \frac{n}{n_{\mathrm{rb}(s)}} .
\]
This ensures that the contributions of presences and both types of pseudo-absences are balanced across species.

Because these ratios can become very large under severe class imbalance, we additionally consider using the square root of the weights to stabilize training, a common practice in machine learning.

An additional advantage of the $\mathcal{L}_{\text{optimized}}$ loss lies in its generality, as several existing losses can be recovered as special cases. In particular, $\mathcal{L}_{\text{baseline}}$ corresponds to $\mathcal{L}_{\text{optimized}}$ with $\lambda_1 = 1$, $\lambda_2 = 0.5$, and uniform weights $w_{\mathrm{p}(s)} = w_{\mathrm{tgb}(s)} = w_{\mathrm{rb}(s)} = 1$.

Beyond this principled weighting scheme, we also consider alternative strategies for handling class imbalance. Specifically, we evaluate an approach that tunes $\lambda_1$ while keeping all species-specific weights equal to $1$. In this setting, $\lambda_1$ is treated as a hyperparameter and is swept over a range of values (from $1$ to $128$) to identify the optimal setting. Finally, we evaluate a hybrid approach that combines both strategies, i.e., tuning $\lambda_1$ in conjunction with species-specific weighting.

\subsection{Order of design choices}
\label{sec:appendix_ablation_order}

The sensitivity analyses in the main text are conducted sequentially: each design choice is optimized in turn, and the best-performing configuration is fixed before proceeding to the next. We consider design choices in the order of the modeling pipeline, from data-related to model-related components: aggregation, subsampling, architecture, and finally loss. This ordering reflects the fact that training data must be finalized before meaningful comparisons between model configurations can be made. It also ensures that each subsequent choice is evaluated on the dataset ultimately used to train the final model. In contrast, optimizing the architecture or loss before refining the training data could lead to selecting configurations specific to an intermediate dataset that become suboptimal once the data processing pipeline is updated.

\section{Additional results}
\label{sec:appendix_results}

\subsection{Fine-grained performance heatmaps}
\label{sec:appendix_finegrained_performance_heatmap}

\begin{figure}[ht]
     \centering
     \includegraphics[width=\textwidth]{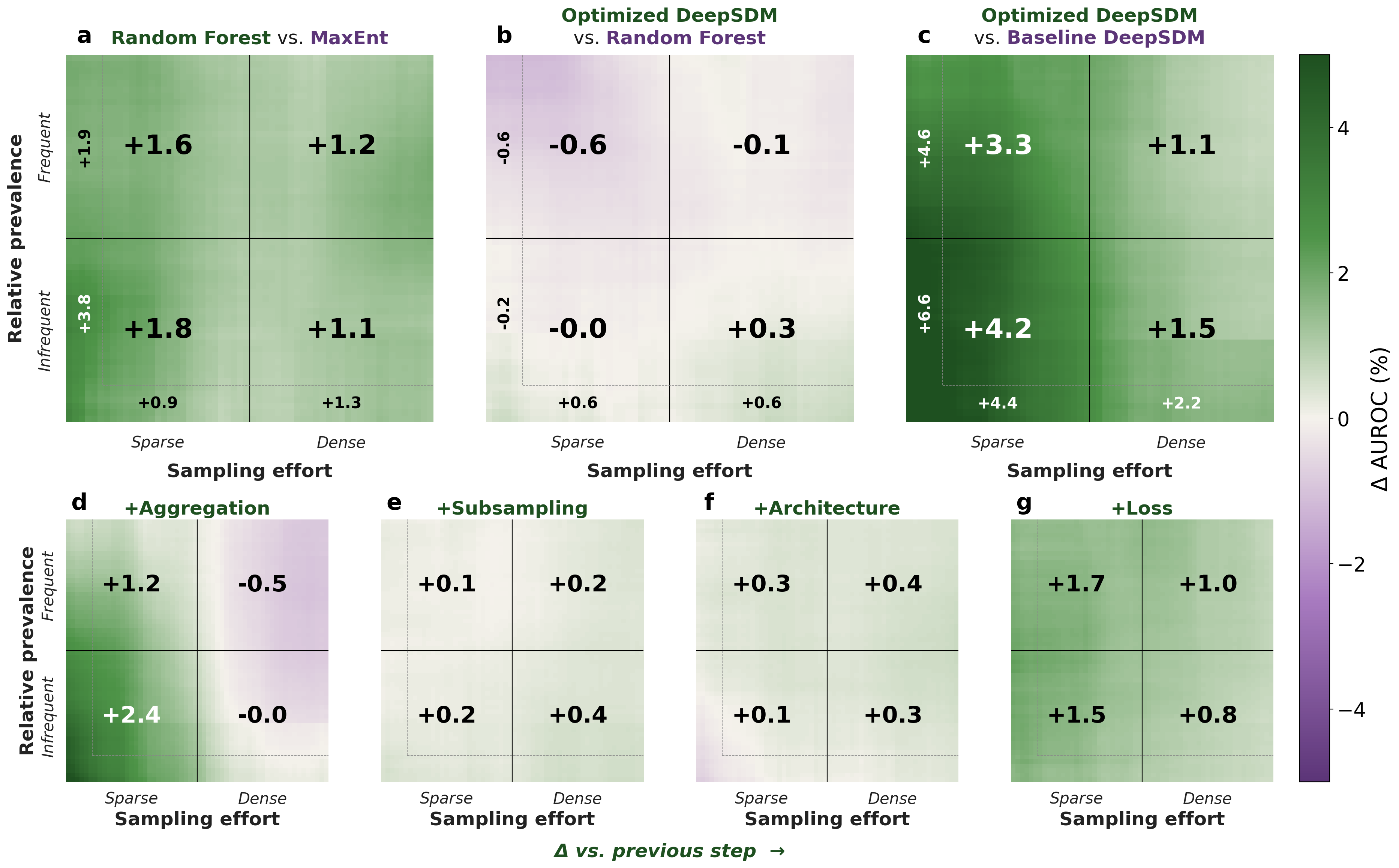}
     \caption{\textbf{Fine-grained performance differences ($\Delta$AUROC, \%) across the property space.} Same panel layout as Figure~\ref{fig:quadrants}, but computed on a smoothed rank grid (window size 2000, step 50) rather than eight discrete regions. Panels \textbf{(a--c)} show pairwise model comparisons; panels \textbf{(d--g)} show the incremental effect of each design choice.}
     \label{fig:delta_heatmap_fine}
\end{figure}

Figure~\ref{fig:delta_heatmap_fine} shows the same comparisons as Figure~\ref{fig:quadrants}, but at higher granularity using a sliding-window average over ranked sampling effort and relative prevalence. This fine-grained view confirms that the coarse quadrants faithfully capture the dominant performance gradients, with no major dynamics lost through the discretization.

\subsection{AUPRG results}
\label{sec:appendix_auprg}
Table~\ref{tab:main_results} reports AUROC, which measures ranking quality across all thresholds. Following the evaluation procedure of \citet{valavi2022predictive}, we additionally evaluate all models using the Area Under the Precision--Recall Gain curve (AUPRG; \citealp{flach2015precision}), which emphasizes the precision--recall trade-off for rare positives. AUPRG re-weights precision and recall so that the baseline corresponding to a random classifier is zero rather than prevalence-dependent, making it better suited for comparing performance across species with widely varying prevalence. Unlike AUROC, however, AUPRG is unbounded below, and a small number of species scoring far below the random baseline dominates the macro-average: for the Random Forest, the worst species reaches $-71.6$ and the lowest percentile $-2.8$, against a median of $0.87$. We therefore summarize AUPRG by the median across species; AUROC, being bounded, is summarized by the mean as before.
\begin{table}
\caption{\textbf{Models performance comparison on \DATASETNAME.} Overall and group-level AUPRG ($\%$) are reported as median\,±\,standard deviation over 5 random seeds.}
\centering
\renewcommand{\arraystretch}{1.25}
\setlength{\tabcolsep}{6pt}
\small
\begin{tabular}{@{}l c cc cc c@{}}
\toprule
& \multicolumn{5}{c}{\textbf{AUPRG ($\uparrow$)}} & \\
\cmidrule(lr){2-6}
& & \multicolumn{2}{c}{\textbf{Sparse}} 
& \multicolumn{2}{c}{\textbf{Dense}} & \\
\cmidrule(lr){3-4} \cmidrule(lr){5-6}
\textbf{Model / Configuration} 
& \textbf{Overall} 
& \textbf{Infrequent} & \textbf{Frequent} 
& \textbf{Infrequent} & \textbf{Frequent}
& \textbf{Time (min)} \\
\midrule
\multicolumn{7}{@{}l}{\textbf{Non-DL models}} \\[2pt]
\quad GLM                & 79.3 \pms{0.2}           & 76.6 & 77.6 & 82.0 & 80.3 & 1667\pms{25}   \\
\quad BRT                & 84.5 \pms{0.2}           & 83.9 & 82.1 & 87.7 & 84.0 & 1152\pms{29}   \\
\quad GAM                & 81.0 \pms{0.2}           & 78.4 & 78.8 & 84.2 & 81.7 & 2073\pms{49}   \\
\quad MaxEnt             & 83.1 \pms{0.2}           & 82.7 & 80.6 & 86.4 & 82.6 & 12383\pms{242} \\
\quad Random Forest      & \textbf{86.9} \pms{0.1}  & \textbf{88.0} & \textbf{84.9} & \underline{89.1} & \textbf{85.4} & 2610\pms{35} \\
\midrule
\multicolumn{7}{@{}l}{\textbf{DeepSDMs}} \\[2pt]
\quad DeepSDM baseline   & 82.7 \pms{0.3}           & 78.2 & 77.9 & 88.5 & 84.8 & 183\pms{6} \\
\quad + Data aggregation & 83.6 \pms{0.2}           & 82.9 & 79.1 & 87.8 & 84.1 & 64\pms{5}  \\
\quad + Subsampling      & 83.5 \pms{0.2}           & 82.0 & 79.3 & 87.9 & 84.2 & 26\pms{2}  \\
\quad + Architecture     & 85.0 \pms{0.3}           & 83.7 & 80.8 & 89.4 & \underline{85.3} & 19 \pms{2} \\
\quad + Loss             & \underline{85.4} \pms{0.2}& \underline{85.2} & \underline{81.7} & \textbf{89.4} & 85.0 & 17\pms{1} \\

\bottomrule
\end{tabular}
\label{tab:results_auprg}
\end{table}

\begin{figure}
     \centering
     \includegraphics[width=\textwidth]{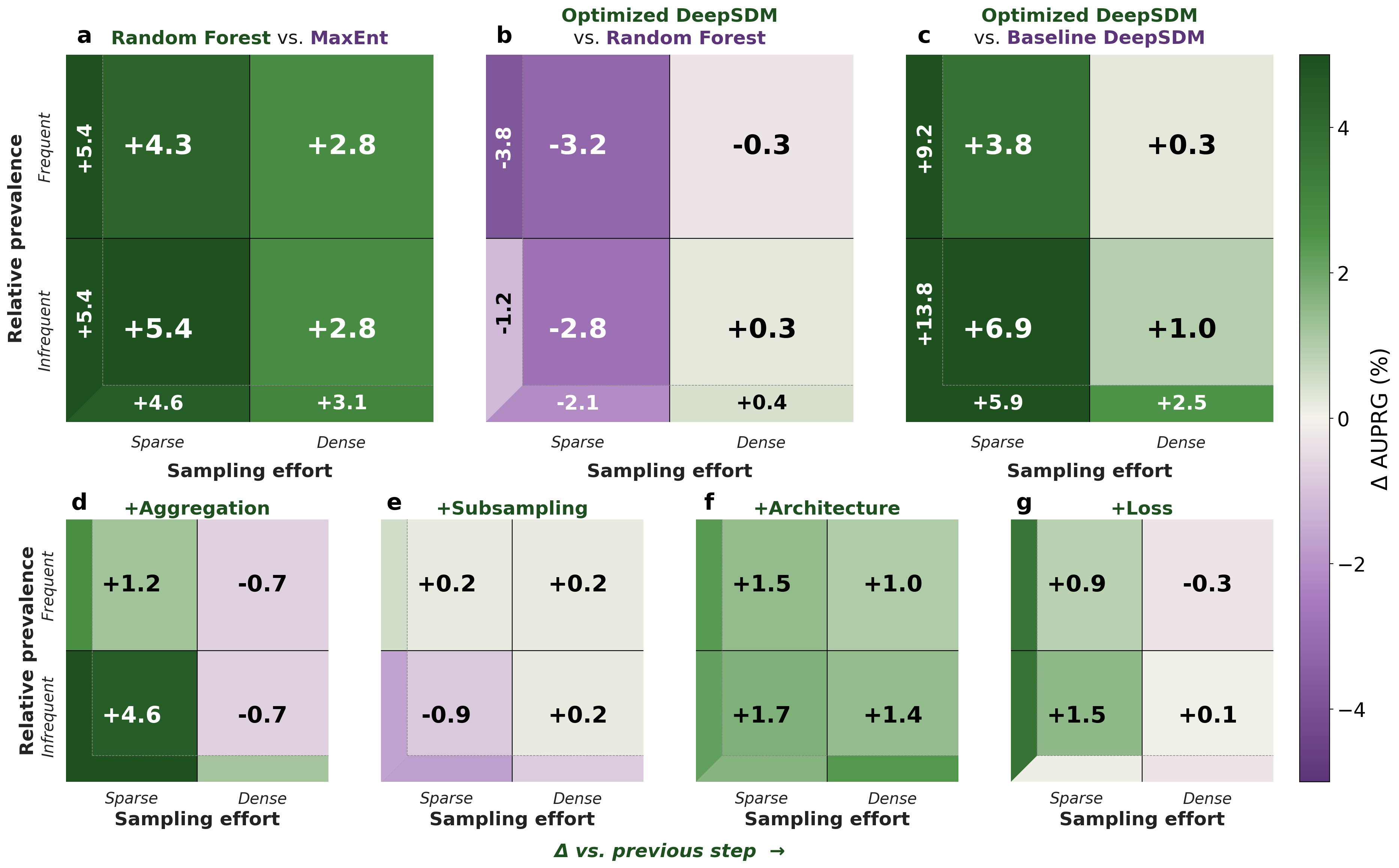}
     \caption{\textbf{Performance differences ($\Delta$AUPRG, \%) across species groups defined by sampling effort and relative prevalence}, mirroring Figure~\ref{fig:quadrants} with AUPRG in place of AUROC.}
     \label{fig:auprg_quadrants}
\end{figure}

The AUPRG results (Table~\ref{tab:results_auprg}) confirm the main AUROC findings while sharpening several contrasts. Random Forest again performs best ($86.9\%$), followed by BRT and MaxEnt; GLM and GAM remain clearly behind, though their order is reversed relative to AUROC, with the GAM ahead ($81.0\%$ against $79.3\%$). The proposed design choices improve the DeepSDM over the baseline (overall $82.7\% \rightarrow 85.4\%$), with the largest gains for the most data-deficient species ($+6.9\%$ for the sparse-infrequent group; Figure~\ref{fig:auprg_quadrants}). Two differences stand out relative to AUROC. First, under AUPRG, Random Forest retains a clearer overall edge over the optimized DeepSDM ($86.9\%$ vs.\ $85.4\%$) than under AUROC, where the two are nearly tied. Second, whereas every design choice yields monotonic improvements in AUROC, subsampling is essentially neutral and the loss reweighting exhibits a trade-off under AUPRG, raising sparse--infrequent performance while slightly lowering the dense-group scores (Figure~\ref{fig:auprg_quadrants}g).

\subsection{Effect of plot size on the evaluation}
\label{sec:appendix_plot_area}
SPlotOpen plots span three orders of magnitude in surveyed area (Section~\ref{sec:appendix_splotopen}), while predictions are made at a 1~km grain. A small plot samples only a small fraction of its grid cell, so a species may be present in the cell yet absent from the plot, which registers as a false positive.

To test whether this affects the comparison between models, we re-evaluate all models with evaluation cells weighted by how much of the cell was actually surveyed. For each cell, we take the total surveyed area as the sum of \texttt{Releve\_area} over the distinct sPlotOpen plots falling in that cell. Of the \num{42268} test cells, \num{26856} have a recorded area (mean \num{530}~m$^2$, median \num{200}~m$^2$); the remaining \num{15412} are assigned the mean recorded area for this analysis. Each per-species AUROC is then recomputed using two weighting strategies: the area itself, and its square root, which dampens the influence of the few very large plots that otherwise dominate.

Both weightings slightly lower absolute scores, but the ranking of all seven models is identical to the unweighted results (Table~\ref{tab:plot_area}). The Random Forest and the optimized DeepSDM remain the two strongest models, separated by $0.1$~\% throughout. The plot-size mismatch therefore affects the absolute level of the reported scores but not the conclusions drawn from comparing models, and we report unweighted results throughout.

\begin{table*}[!t]
  \caption{\textbf{Model performance under area-weighted evaluation.} AUROC (\%) averaged over all species, with each evaluation cell weighted uniformly, by the square root of its surveyed area, or proportionally to it. Cells without a recorded area are assigned the mean recorded area (\num{530}~m$^2$). Scores are computed from the seed-0 run of each model. Down-weighting small plots lowers every score slightly and leaves the ranking unchanged.}
  \label{tab:plot_area}
  \centering
  \small
  \begin{tabular}{@{}lcccccc@{}}
  \toprule
  & \multicolumn{2}{c}{\textbf{Unweighted}} & \multicolumn{2}{c}{\textbf{$\sqrt{\text{Area}}$-weighted}} & \multicolumn{2}{c}{\textbf{Area-weighted}} \\
  \cmidrule(lr){2-3} \cmidrule(lr){4-5} \cmidrule(lr){6-7}
  \textbf{Model} & AUROC & Rank & AUROC & Rank & AUROC & Rank \\
  \midrule
  \multicolumn{7}{@{}l}{\textbf{Single-species SDMs}} \\[2pt]
  \quad GLM               & 84.0 & 5 & 83.8 & 5 & 83.8 & 5 \\
  \quad BRT               & 85.3 & 3 & 85.1 & 3 & 85.1 & 3 \\
  \quad GAM               & 83.7 & 6 & 83.5 & 6 & 83.5 & 6 \\
  \quad MaxEnt            & 84.5 & 4 & 84.3 & 4 & 84.4 & 4 \\
  \quad RF                & \textbf{85.9} & 1 & \textbf{85.8} & 1 & \textbf{85.8} & 1 \\
  \midrule
  \multicolumn{7}{@{}l}{\textbf{Multi-species DeepSDMs}} \\[2pt]
  \quad Baseline          & 83.3 & 7 & 83.1 & 7 & 83.1 & 7 \\
  \quad Optimized DeepSDM & 85.8 & 2 & 85.7 & 2 & 85.7 & 2 \\
  \bottomrule
  \end{tabular}
\end{table*}

\clearpage

\end{document}